\documentclass{article}

\usepackage[preprint]{neurips_2026}
\usepackage{booktabs}
\usepackage{url}
\usepackage{hyperref}

\newcommand{\codeurl}{\url{https://anonymous.4open.science/r/expert-cd-ensemble-3282/}}

\usepackage{amsmath}
\usepackage{amssymb}
\usepackage{algorithm2e}
\SetKw{kwin}{in}
\SetKw{kwand}{and}
\SetKw{kwor}{or}
\SetKw{kwnot}{not}
\usepackage{textpos}

\usepackage{graphicx}

\usepackage{xcolor}         

\usepackage{colortbl}

\usepackage{cleveref}
\crefname{section}{§}{§§}
\Crefname{section}{§}{§§}

\title{Dynamic Expert-Guided Model-Averaging\\for Causal Discovery}

\author{%
  Adrick Tench\\
  Ghent University - imec\\
  \texttt{adrick.tench@ugent.be} \\
  \And
  Thomas Demeester\\
  Ghent University - imec\\
  \texttt{thomas.demeester@ugent.be} \\
}

\begin{document}

\maketitle

\begin{abstract}
    Would-be practitioners of causal discovery face a dizzying array of algorithms without a clear best choice. This abundance of competitive methods makes ensembling a natural strategy for practical applications. At the same time, real-world use cases frequently violate the assumptions on which common causal discovery algorithms are based, forcing reliance on expert knowledge. Inspired by recent work on dynamically requested expert knowledge and large language models (LLMs) as experts, we present a flexible model averaging method that integrates selective expert querying to ensemble a diverse set of causal discovery algorithms. Crucially, we distinguish between edge existence and orientation, enabling the method to leverage the complementary strengths of data-driven discovery and expert input. We further consider the realistic setting of limited access to an imperfect expert, using disagreement among algorithms to query the expert in cases of greater uncertainty. Experiments demonstrate that our method consistently outperforms strong baselines on both clean and noisy data. Code and data are available at \codeurl.
\end{abstract}

\section{Introduction}

Applications across many domains require not just accurate predictive models, but a robust understanding of the underlying causal mechanisms. Probabilistic graphical models such as Bayesian networks (BNs) provide a principled framework for representing causal relationships, offering inherent interpretability and capacity for counterfactual and interventional reasoning \citep{pearl2009causality,koller2009probabilistic}. Causal discovery aims to learn causal relations between variables 
from observational data, revealing the underlying structure of complex systems 
\citep{spirtes2000causation}. 

Randomized controlled trials (RCTs) are the gold standard for determining causal relationships, but conducting RCTs is often costly and difficult, if not impossible. Causal discovery can point the way towards the most useful trials to perform and provide insight in cases where trials are infeasible. However, would-be practitioners of causal discovery face a great diversity of methods, with no clear indicators of which method is the best for their use-case \citep{kitson2023survey}. This, combined with the realities of limited, noisy data, forces heavy reliance on expert knowledge in practice \citep{vonk2023disentangling, makela2022incorporating}.

We therefore seek to provide a general-purpose method for ensembling in causal discovery that can flexibly incorporate a diverse array of learning algorithms with expert domain knowledge. Our primary contributions are threefold:
\begin{itemize}
    \item We provide a model averaging strategy that uses dynamically requested expert knowledge to mediate discrepancies between models. To our knowledge, our method is the first to explore the use of expert knowledge for model averaging or ensembling in causal discovery.
    \item We treat the questions of edge existence and edge orientation separately and selectively query the expert in cases of greater disagreement between models, seeking to maximize the value of limited access to an imperfect expert.
    \item We evaluate the performance of our method against strong baselines on well-known, real-world graphs, using both clean and noisy data to better reflect the challenges of real-world applications of causal discovery.
\end{itemize}

\begin{figure}[h]
    \centering
    \includegraphics[width=\textwidth]{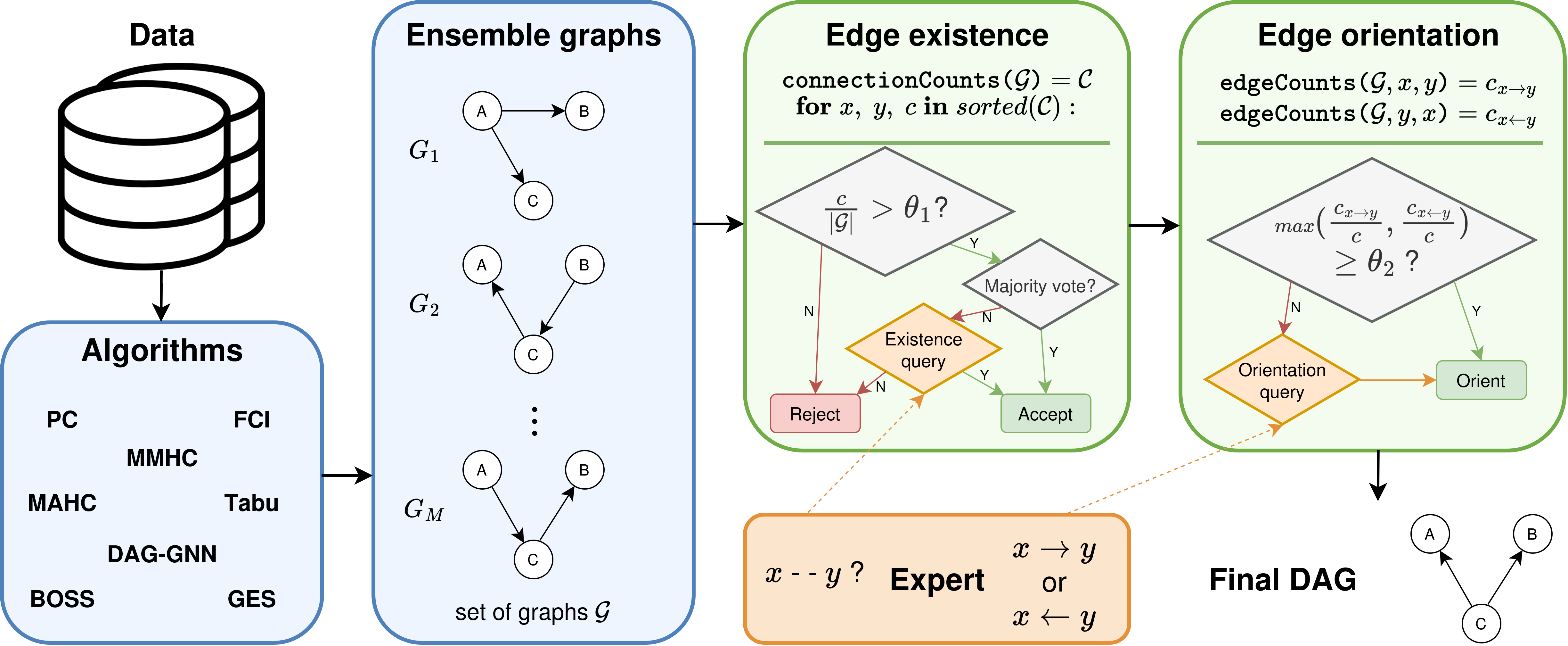}
    \caption{Overview of our method. We run a set of causal discovery algorithms on the input data, producing a set of graphs $\mathcal{G}$. We greedily add edges by order of most frequent appearance, irrespective of orientation. If the percentage of models containing a direct edge exceeds a predefined threshold $\theta_1$, the edge is considered for addition. If the percentage passes a majority vote, the edge is added; otherwise, the expert is queried to validate the existence of the edge. For accepted edges, orientation is then determined. If the percentage of models containing a particular orientation (out of the models containing a direct edge) exceeds a predefined threshold $\theta_2$, that orientation is added. Otherwise, the expert is queried to determine orientation. The expert is queried up to a query budget $B$, after which decisions default to majority voting. Enforcement of the query budget and acyclicity constraint is not shown to avoid cluttering the figure.}
    \label{fig:overview}
\end{figure}

Figure~\ref{fig:overview} illustrates our proposed approach, further detailed in the subsequent sections.

\section{Related Work}\label{sec:related}

\subsection{Assumptions and data noise}
The theoretical bases for many causal discovery algorithms rely on assumptions that are frequently impossible to verify in practice, such as causal sufficiency (the absence of latent confounders). A plethora of algorithms exist that attempt to address different possible assumption violations. For example, the well-known FCI algorithm \citep{spirtes2000causation} works in the presence of latent confounders and selection bias \citep{spirtes2013causal}; the SAT-based approach of \citet{hyttinen2013discovering} handles latent confounders and cyclicity; and the hill-climbing based approach of \citet{liu2022greedy} is shown to improve performance in the face of missing data, including data Missing Not At Random. However, such approaches typically only address one or two assumptions while leaving others untouched. Furthermore, analysis of the performance of well-known causal discovery algorithms under data noise that violates various common assumptions reveals inconsistencies in algorithm performance across metrics and noise types \citep{constantinou2021large}.
As no algorithm consistently outperforms all others on real-world data, we opt for an ensembling approach that leverages the potential advantages of a multitude of algorithms.

\subsection{Ensembling in causal discovery}
The performance of individual causal discovery algorithms can vary widely and unpredictably \citep{kitson2023survey}, making the improved robustness provided by ensembling \citep{zhou2025ensemble} a natural choice for applying causal discovery in practical settings. However, to our knowledge, little research has been done exploring the best methods for performing ensembling or model averaging in causal discovery. \citet{constantinou2019bayesys} provides a method for model averaging that greedily adds edges by order of the most counts, avoiding cycles as the graph is constructed. \citet{aslani2023ensemble} treat ensembling as an optimization problem, using a distance measure between graphs to maximize the marginal contribution of the edges in the final graph. \citet{khatibi2024alcm} present their strongest results from a weighted ensemble of existing causal discovery algorithms. However, their approach relies on access to unrealistic metadata about the ground-truth causal graph that would not be available in realistic settings, and we therefore do not consider this scenario.

\subsection{Expert knowledge}
Prior work has typically focused on including expert knowledge as prior constraints \citep{constantinou2023impact}. A recent contribution from \citet{kitson2025causal} explored using dynamically-requested expert knowledge in the score-based tabu algorithm. Unlike our approach, their method is specifically intended for use with a human expert, whereas our approach permits a human or large language model (LLM) 
expert, and evaluation largely focused on a setting with clean data and a perfect (simulated) expert.

There has also been much interest in using LLMs as experts to enable causal discovery \citep{liu2025large}. \citet{vashishtha2025causal} present a strong case for restricting LLMs to answering causal ordering questions, and provide a method for prompting LLMs to obtain a causal order over a whole graph, to be used as prior constraints. \citet{khatibi2024alcm} introduce a general framework for using an LLM to refine the output of a causal discovery algorithm. In line with the dynamic knowledge method of \citet{kitson2025causal}, \citet{tao2025active} use active requests to an LLM to augment continuous optimization algorithms (e.g. \citet{zheng2018dags}'s NOTEARS).

Inspired by these contributions, we explore the use of imperfect experts (such as LLMs) to dynamically answer causal ordering questions, improving model averaging across an ensemble of varied causal discovery algorithms. Unlike previous methods, we explicitly use expert knowledge to mediate the ensembling process.

\section{Problem Setup}\label{sec:setup}

\subsection{Definitions}

Let $\mathcal{V} = \{V_1, \dots, V_n\}$ denote a set of variables, and let $G^\star$ denote the unknown ground-truth causal graph over $\mathcal{V}$, which we assume to be a directed acyclic graph (DAG). A partially directed acyclic graph (PDAG) over~$\mathcal{V}$ is a graph containing directed or undirected edges with no directed cycles.

We are given a set of $M$ causal discovery algorithms, each producing a PDAG 
over $\mathcal{V}$.  We collect their outputs in the set
\[\mathcal{G} = \{G_1, \dots, G_M\} \subseteq \mathcal{G}_{\text{PDAG}}(\mathcal{V})\]
which forms the input to our graph averaging procedure.
%
For any unordered pair of variables $(x,y)$, we define:

\begin{itemize}
    \item \textbf{Edge existence:} $Z_{xy}^\star \in \{0,1\}$, where $Z_{xy}^\star = 1$ indicates the presence of an edge between $V_x$ and $V_y$ in $G^\star$.
    
    \item \textbf{Edge orientation:} $O_{xy}^\star \in \mathcal{O}$, where $\mathcal{O} = \{\to,\; \gets\}$ denotes edge direction, defined only when $Z_{xy}^\star = 1$.
\end{itemize}

Each learned graph $G_m$ provides corresponding values $Z_{xy}^{(m)}$ and $O_{xy}^{(m)}$. As graphs in $G_m$ may contain undirected edges, $O_{xy}^{(m)}$ may take values in $\mathcal{\hat{O}} = \{\to,\; \gets,\; -\}$.

We additionally have access to an imperfect expert that can be queried about the existence and orientation of specific edges:

\begin{itemize}
    \item \textbf{Expert edge existence:} $Z_{xy}^e \in \{0,1\}$.
    \item \textbf{Expert edge orientation:} $O_{xy}^e\in \mathcal{O}$.
\end{itemize}


\subsection{Assumptions}

Standard approaches to causal discovery rely on strong assumptions such as Causal Markov, faithfulness, and observability, which we discuss in Appendix \ref{app:standard_assumptions}. However, rather than resting our method on these formal assumptions that are often violated in practical settings \citep{montagna2023assumption}, we focus on weaker assumptions which we can reasonably expect to hold in practice, motivating an expert-guided ensembling approach for causal discovery. Given these assumptions and the deliberate involvement of an (imperfect) expert, we correspondingly rest our method on empirical validation rather than formal guarantees of performance in the infinite data limit.

\paragraph{Assumption 1 \normalfont(Algorithm competence)\textbf{.}} 
Algorithms perform better than random guessing on edge existence:
\[
\mathbb{P}\left(Z_{xy}^{(m)} = Z_{xy}^\star\right) > \frac{1}{2}, 
\quad \forall\;m \in \{1,\dots,M\}.
\]
Intuitively, this assumption suggests that adding algorithms to the ensemble will increase its performance, approaching perfect performance with the addition of further algorithms \citep{de1785essais}. Appendix \ref{app:algorithm_competence} discusses this assumption further, exploring why simply adding additional (competent) algorithms to the ensemble cannot be expected to approach perfect performance in practice, and how the use of an expert may make our method robust to minor violations of this assumption. 

\paragraph{Assumption 2 \normalfont(Algorithm diversity)\textbf{.}} 
The algorithms exhibit disagreement. That is, there exist pairs $(i,j)$ such that:
\[
\exists\, m \neq m' \text{ with } Z_{xy}^{(m)} \neq Z_{xy}^{(m')} \quad \text{or} \quad O_{xy}^{(m)} \neq O_{xy}^{(m')}.
\]
This assumption is trivially necessary for the ensembling process to provide any value.

\paragraph{Assumption 3 \normalfont(Expert competence)\textbf{.}}
The expert performs better than random guessing:
\[
\mathbb{P}(Z_{xy}^e= Z_{x\dots y}^\star) = p_e^{\text{exist}} > \frac{1}{2}\quad\text{and}\quad
\mathbb{P}(O_{xy}^e= O_{xy}^\star \mid Z_{xy}^\star = 1) = p_e^{\text{orient}} > \frac{1}{2},
\]
where $Z_{x\dots y}^\star=1$ denotes that the true graph contains some chain $x \to \ldots \to y$, which may or may not be a direct edge.\footnote{As argued by \citet{vashishtha2025causal}, determining the presence of a direct edge in a causal graph generally requires the consideration of every node in the graph, making existence queries potentially risky. We therefore focus on the potential impact of causal ordering questions, assuming that the expert will typically only reject an edge in the event that there is no plausible causal link, direct or indirect.} Appendix \ref{app:expert_correctness} explores how our method behaves with varying degrees of expert correctness, providing empirical evidence that better-than-random expert performance is a necessary condition to outperform an ensembling method that does not use an expert.

\paragraph{Assumption 4 \normalfont(Expert cost)\textbf{.}} 
The expert will answer only a limited number of queries. Let $\mathcal{Q}$ denote the set of queries performed and $B$ a fixed query budget. Then:
\[
|\mathcal{Q}| \leq B.
\]
This reflects typical situations in practice, with time and/or budget constraints limiting access to human or LLM experts.

\paragraph{Assumption 5 \normalfont(Query value)\textbf{.}} 
The expected value of querying an edge is positively correlated with the level of disagreement among algorithms. Informally, edges for which the set $\{G_m\}$ exhibits greater disagreement are more likely to benefit from expert input. We provide some validation for this assumption in Appendix \ref{app:assumption_validation}, finding that querying the expert without regard to disagreement performs worse than our strategy which prioritizes resolving disagreement.

\subsection{Objective}

Given a set of learned graphs $\mathcal{G} = \{G_1, \dots, G_M\}$ and a limited query budget $B$ for accessing an imperfect expert, the goal is to learn a graph $\hat{G}$ that is as close as possible to the unknown ground-truth graph $G^\star$.

Formally, we aim to solve:
\[
\max_{\hat{G},\, \mathcal{Q}} \; \mathbb{E}\left[\mathrm{Score}(\hat{G}, G^\star)\right]
\quad \text{subject to} \quad |\mathcal{Q}| \leq B,
\]
where $\mathrm{Score}(\cdot, \cdot)$ measures similarity (e.g., F1 score or another suitable metric).

As in practice, $G^\star$ is unknown, the objective cannot be optimized directly. We therefore propose to approximate the solution with a greedy procedure that prioritizes high-confidence edges and selectively queries uncertain ones. 

\section{Methods}\label{sec:methods}

\subsection{Separation of edge existence and orientation}

We choose to treat the questions of edge existence and orientation separately, motivated by two key observations:

\paragraph{Observation 1 \normalfont(Algorithm orientation ambiguity)\textbf{.}} In general, even under ideal conditions, it is only possible to learn a Completed Partially Directed Acyclic Graph (CPDAG) from observational data, representing a Markov Equivalence Class \citep{spirtes2000causation}. This CPDAG typically contains some undirected edges for which the orientation cannot be determined from data alone. Furthermore, given an ensemble of diverse algorithms working with limited data, it is reasonable to expect cases where multiple algorithms identify that an edge exists while disagreeing on its orientation.

\paragraph{Observation 2 \normalfont(Expert orientation accuracy)\textbf{.}} Prior research suggests that LLMs perform quite well at determining the causal direction between pairs of variables \citep{kiciman2023causal}, but can struggle with distinguishing the existence of a causal relationship from mere correlation \citep{carro2024large}. Human experts are similarly subject to biases that result in conflating correlation with causation \citep{msetfi2013context, matute2015illusions}. Therefore, we only query the expert for edges detected by the ensemble, not to introduce new edges.

Treating edge existence and orientation separately allows our method to rely more on data-driven algorithms for detecting edge existence and experts for determining their orientation. Furthermore, it allows for distinct, targeted strategies when soliciting expert opinions on existence and orientation.

\subsection{Expert model averaging}\label{sec:ema_method}

We describe our dynamic expert-guided model averaging strategy below. Detailed pseudocode is provided in Appendix~\ref{app:pseudocode}.

Inspired by the model averaging strategy of \citet{constantinou2019bayesys}, our method greedily adds edges that avoid creating cycles by order of most occurrence in the component models $\mathcal{G}$. However, crucially, our method treats the existence of a direct connection separately from the orientation of that connection: when determining the order in which edges will be added, any direct edge between two nodes $x$ and $y$ (i.e. $x\rightarrow y, y \rightarrow x, x - y$) is weighed equally towards establishing that there is some direct connection, irrespective of orientation.


We use a predetermined edge threshold $\theta_1$ that determines what percentage of component models $\mathcal{G}$ must contain a connection between $x$ and $y$ for an edge to be added to $\hat{G}$. When the percentage is above $\theta_1$ but below 0.5 (a majority vote), we actively query the expert $e$ to potentially reject the presence of an edge between $x$ and $y$.

In the event that only one directed orientation ($x\rightarrow y$ or $y \rightarrow x$) is a valid (non-cycle inducing) addition, that edge is added to $\hat{G}$. When both directed orientations are valid, we first check if the percentage of models with a particular orientation ($O_{xy}^{(m)} =\,\to$ or $\gets$) 
out of the models containing a direct connection ($Z_{xy}^{(m)}=1$) 
falls above a predetermined orientation threshold $\theta_2$. If so, that orientation is added to $\hat{G}$. Otherwise, the expert $e$ is actively queried to determine the orientation.

Finally, we include a query budget $B$, indicating how many queries can be posed to the expert. When the budget is exhausted, the ensembling defaults to majority voting.

\subsection{Simulated expert}\label{sec:sim_expert}

We use a simulated expert in our experiments. Based on the ground-truth model $G^\star$, the expert provides correct answers to existence and orientation queries in some fixed percentage of cases $p_e^{\text{exist}}$ and $p_e^{\text{orient}}$, respectively. As discussed in Assumption 3, we assume the expert answers existence queries on the basis of potentially indirect causal ordering. Our simulated expert therefore provides answers based on presence of some chain $x \to \ldots \to y$ rather than merely a direct edge $x \rightarrow y$. From the perspective of graphical accuracy, a correct causal ordering answer from our simulated expert may therefore not result in a correct edge in the final graph (i.e. if there is a chain but no edge between $x$ and $y$ in $G^\star$, when answering ``correctly''  our expert will not block an edge between them). We choose to take this approach to better reflect the limitations of eliciting accurate expert knowledge in practice \citep{vashishtha2025causal}.  We analyze how our approach performs when using a simulated expert with access to knowledge of direct edges in Appendix \ref{app:direct_edge_expert}, finding as expected that such an expert provides better precision but identical recall. Finally, we note that when orientation is queried for two variables with no causal relation in $G^\star$, an arbitrary orientation is returned.

\section{Experiments}

\subsection{Setup}


\paragraph{Datasets:} We use six networks taken from the Bayesys repository \citep{constantinou2020bayesys}, containing real-world BNs from various application domains, described in Table~\ref{tab:networks}. We use the corresponding clean data and the noisy data generated in \citet{constantinou2021large}, with sample sizes $n \in \{100, 1000, 10000, 100000\}$. We also include the SimSUM network \citep{rabaey2025simsum}, with noisy data generated following the same procedure as \citet{constantinou2021large}. To mitigate the impact of variable ordering on results \citep{kitson2024impact}, we repeat each experiment over twenty random orderings of the dataset columns.

\begin{table}[ht]
    \caption{Properties of benchmark networks}
    \label{tab:networks}
    \centering
    \scalebox{0.8}{
    \begin{tabular}{lllll}
        \toprule
        \textbf{Network} & \textbf{Nodes} & \textbf{Edges} & \textbf{Parameters} & \textbf{Domain} \\
        \midrule
        Asia        & 8   & 8   & 18      & Reasons causing dyspnea \\
        Sports      & 9   & 15  & 1,049   & Football match outcomes \\
        Property    & 27  & 31  & 3,056   & Property investment \\
        Alarm       & 37  & 46  & 509     & ALARM monitoring system \\
        Formed      & 88  & 138 & 912     & Forensic / criminal psychology \\
        Pathfinder  & 109 & 195 & 71,890  & Diagnosis of lymph-node diseases \\
        SimSUM      & 16  & 35  & 3,368    & Respiratory diseases \\
        \bottomrule
    \end{tabular}}
\end{table}

\paragraph{Algorithms and baselines:}
We use a diverse array of causal discovery algorithms in our ensemble: the constraint-based PC-Stable \citep{colombo2014order} and FCI \citep{spirtes2000causation}; the score-based Tabu \citep{bouckaert1995bayesian}, GES \citep{chickering2002learning}, MAHC \citep{constantinou2022effective}, and BOSS \citep{andrews2023fast}; the hybrid MMHC \citep{tsamardinos2006max}; and the continuous optimization approach DAG-GNN \citep{yu2019dag}, chosen for its handling of discrete variables. We also compare to two existing ensembling approaches from \citet{constantinou2019bayesys} and \citet{aslani2023ensemble}. We include Tabu-AL \citep{kitson2025causal} as a baseline using dynamically requested expert knowledge, using the simulated expert described in Section \ref{sec:sim_expert}. Hyperparameter choices and implementation details for the algorithms are contained in Appendix \ref{app:h_imp}. Finally, we include an Oracle baseline that selects the best individual method of the ensemble for a given network, sample size, noise setting, and metric. We note that this baseline forms an unreasonably high bar for comparison, as in practice the best individual method would not be known ahead of time, and furthermore may vary by metric even with other settings fixed.

\paragraph{Metrics:}
We report several evaluation metrics, namely the Balanced Scoring Function (BSF) \citep{constantinou2019evaluating}, F1-score, precision score, recall score, Structural Hamming Distance (SHD), and Structural Intervention Distance (SID) \citep{peters2013structural}. In all cases (save the SID, which is computed differently), we treat a reversed or undirected edge (i.e. $x \leftarrow y$ or $x - y$ when the ground truth is $x \rightarrow y$) as a partial match (0.5 of a false negative and 0.5 of a true positive), in line with \citet{constantinou2019evaluating}, to reflect the relative value of discovering a true dependency without discovering its correct orientation.\footnote{As in \citet{constantinou2022effective}, in the event of a latent confounder under noisy data, only a bidirected edge counts as a full match, while any other edge counts as a partial match. This means that of our evaluated algorithms, only FCI can potentially achieve a perfect score in the noisy data case.}

\subsection{Comparison results}

We provide our main results using a simulated expert that provides correct responses to existence queries in $p_e^{\text{exist}} = 70\%$ of cases and orientation queries in $p_e^{\text{orient}} = 90\%$ of cases. Appendix \ref{app:llm_results} shows that LLM performance on a subset of the benchmark networks is in-line with these values, supporting this choice.
We set our hyperparameters to an edge threshold $\theta_1$ of $0.0$, an orientation threshold $\theta_2$ of $0.9$, and a query budget of $2N$, where $N$ is the number of nodes. These values are chosen to optimize for the BSF and F1 scores, as discussed in Appendix \ref{app:hyperparameter_choices}.

Under these settings, we analyze the performance of our method in Figure \ref{fig:delta_comparison}. Given the large number of combinations of networks, sample sizes, and algorithms, it is impossible to provide direct comparisons for all settings in the main text. 
For compactness and to assist overall comparison, Figure \ref{fig:delta_comparison} provides the metric deltas between our method and the baselines, averaged across all networks, sample sizes, and seeds, for both clean and noisy data. Tables \ref{tab:sig_clean} and \ref{tab:sig_noisy} analyze the statistical significance of improvements of our method over baselines, on clean and noisy data respectively, demonstrating consistent improvements of our method in the considered settings. Full result tables by network and sample size are available in Appendix \ref{app:complete_results}.

Some results are excluded from calculations due to either an invalid precision score (in the case that a method returned a fully sparse graph) or a missing result (due to a timeout or exception). We acknowledge that this may bias results, likely in favor of the algorithm that provided an exceptional result (as such exceptions typically occur in more difficult settings). As our method yields no such exceptional results, we therefore expect any bias in the reported results to be in favor of the baselines that do. 
More detail is provided in Appendix \ref{app:excluded_results}.

\begin{figure}[!ht]
    \centering
    \includegraphics[width=\textwidth]{images/delta_comparison.png}
    \caption{The mean delta in graphical accuracy metrics between baselines and our method, showing broad improvements on all metrics save precision. $*$ indicates some missing results for an algorithm due to timeouts ($>$6h) or exceptions. $\dagger$ indicates some invalid results for an algorithm (only applicable to precision). Missing and invalid results are excluded from the calculation.}
    \label{fig:delta_comparison}
\end{figure}
\vspace{-4pt} 

\begin{table}
    \caption{The fraction of clean data settings that our method significantly improves over baselines on various metrics, as determined by a one-sided Wilcoxon signed-rank test over 20 seeds ($p < 0.05$), demonstrating improvement in most settings. A * further indicates that our method improves significantly when considering all settings together.}
    \label{tab:sig_clean}
    \centering
    \resizebox{\textwidth}{!}{
    \begin{tabular}{lccccccc}
    \toprule
    Algorithm & BSF & F1 & Precision & Recall & SHD & SID (lower bound) & SID (upper bound) \\
    \midrule
    PC-Stable & 92\% (24/26)* & 69\% (18/26)* & 4\% (1/26) & 100\% (26/26)* & 50\% (13/26)* & 42\% (11/26) & 85\% (22/26)* \\
    FCI & 96\% (25/26)* & 81\% (21/26)* & 0\% (0/26) & 100\% (26/26)* & 50\% (13/26)* & 31\% (8/26) & 100\% (26/26)* \\
    MMHC & 100\% (28/28)* & 86\% (24/28)* & 11\% (3/28) & 100\% (28/28)* & 54\% (15/28)* & 89\% (25/28)* & 89\% (25/28)* \\
    BOSS & 71\% (17/24)* & 79\% (19/24)* & 58\% (14/24)* & 62\% (15/24)* & 71\% (17/24)* & 33\% (8/24) & 83\% (20/24)* \\
    Tabu & 86\% (24/28)* & 82\% (23/28)* & 50\% (14/28)* & 86\% (24/28)* & 71\% (20/28)* & 86\% (24/28)* & 86\% (24/28)* \\
    GES & 100\% (28/28)* & 96\% (27/28)* & 46\% (13/28)* & 100\% (28/28)* & 82\% (23/28)* & 96\% (27/28)* & 96\% (27/28)* \\
    MAHC & 96\% (25/26)* & 88\% (23/26)* & 23\% (6/26) & 92\% (24/26)* & 77\% (20/26)* & 100\% (26/26)* & 100\% (26/26)* \\
    DAG-GNN & 91\% (20/22)* & 95\% (21/22)* & 73\% (16/22) & 86\% (19/22)* & 91\% (20/22)* & 86\% (19/22)* & 86\% (19/22)* \\
    Tabu-AL & 88\% (21/24)* & 83\% (20/24)* & 75\% (18/24)* & 88\% (21/24)* & 92\% (22/24)* & 79\% (19/24)* & 79\% (19/24)* \\
    Bayesys Model Avg & 61\% (17/28)* & 96\% (27/28)* & 100\% (28/28)* & 36\% (10/28) & 100\% (28/28)* & 61\% (17/28)* & 79\% (22/28)* \\
    Aslani \& Mohebbi & 75\% (15/20)* & 90\% (18/20)* & 100\% (20/20)* & 35\% (7/20) & 100\% (20/20)* & 80\% (16/20)* & 80\% (16/20)* \\
    Oracle & 50\% (14/28)* & 57\% (16/28)* & 0\% (0/28) & 54\% (15/28)* & 18\% (5/28) & 7\% (2/28) & 57\% (16/28)* \\
    \bottomrule
    \end{tabular}
    }
\end{table}

\begin{table}
    \caption{As Table \ref{tab:sig_clean}, for noisy data settings.}
    \label{tab:sig_noisy}
    \centering
    \resizebox{\textwidth}{!}{
    \begin{tabular}{lccccccc}
    \toprule
    Algorithm & BSF & F1 & Precision & Recall & Relative SHD & Relative SID (lower bound) & Relative SID (upper bound) \\
    \midrule
    PC-Stable & 96\% (26/27)* & 93\% (25/27)* & 22\% (6/27) & 100\% (27/27)* & 59\% (16/27)* & 48\% (13/27)* & 89\% (24/27)* \\
    FCI & 96\% (24/25)* & 100\% (25/25)* & 56\% (14/25)* & 100\% (25/25)* & 80\% (20/25)* & 28\% (7/25) & 92\% (23/25)* \\
    MMHC & 93\% (26/28)* & 89\% (25/28)* & 4\% (1/28) & 100\% (28/28)* & 57\% (16/28)* & 86\% (24/28)* & 86\% (24/28)* \\
    BOSS & 92\% (22/24)* & 92\% (22/24)* & 46\% (11/24)* & 100\% (24/24)* & 79\% (19/24)* & 50\% (12/24)* & 88\% (21/24)* \\
    Tabu & 93\% (26/28)* & 89\% (25/28)* & 25\% (7/28) & 100\% (28/28)* & 64\% (18/28)* & 93\% (26/28)* & 93\% (26/28)* \\
    GES & 93\% (26/28)* & 96\% (27/28)* & 25\% (7/28) & 100\% (28/28)* & 79\% (22/28)* & 93\% (26/28)* & 93\% (26/28)* \\
    MAHC & 93\% (26/28)* & 96\% (27/28)* & 25\% (7/28) & 100\% (28/28)* & 68\% (19/28)* & 93\% (26/28)* & 93\% (26/28)* \\
    DAG-GNN & 82\% (18/22)* & 100\% (22/22)* & 45\% (10/22) & 91\% (20/22)* & 77\% (17/22)* & 82\% (18/22)* & 82\% (18/22)* \\
    Tabu-AL & 96\% (25/26)* & 92\% (24/26)* & 27\% (7/26) & 100\% (26/26)* & 65\% (17/26)* & 77\% (20/26)* & 77\% (20/26)* \\
    Bayesys Model Avg & 71\% (20/28)* & 71\% (20/28)* & 96\% (27/28)* & 39\% (11/28)* & 96\% (27/28)* & 64\% (18/28)* & 82\% (23/28)* \\
    Aslani \& Mohebbi & 80\% (16/20)* & 80\% (16/20)* & 95\% (19/20)* & 50\% (10/20)* & 90\% (18/20)* & 90\% (18/20)* & 90\% (18/20)* \\
    Oracle & 64\% (18/28)* & 68\% (19/28)* & 0\% (0/28) & 93\% (26/28)* & 29\% (8/28) & 18\% (5/28) & 68\% (19/28)* \\
    \bottomrule
    \end{tabular}
    }
\end{table}

Our results show broad improvements in BSF and F1 score, reliably improving over baseline methods by several percentage points on both clean and noisy data. The relative SHD and SID also typically improve by wide margins, with the exception of only the unrealistic Oracle baseline. Our method further provides a comparable recall to other ensembling methods with much higher precision. Finally, our method widely improves over Tabu-AL on all metrics (save precision with noisy data) by at least 11\% on average, demonstrating the value of using dynamic expert knowledge for ensembling compared to augmenting an individual algorithm. Tables \ref{tab:sig_clean} and \ref{tab:sig_noisy} show these improvements are statistically significant in most settings.

While individual algorithms often have higher precision, this is balanced by the greatly improved recall of our method. A higher edge threshold $\theta_1$ can easily be used to achieve higher precision at the cost of lower recall, as shown in Appendix \ref{app:edge_threshold}.

\subsection{Query budget}

Table \ref{tab:query_budget} displays the performance of our method across different values of the query budget $B$. Generally, as $B$ increases, recall increases while precision decreases. As our method defaults to majority voting when the query budget is exhausted, this indicates that simple majority voting provides higher precision but lower recall than our method, which incorporates an imperfect expert. Overall, we find that a moderate query budget of $2N$ provides the best balance of BSF, F1, recall, and precision scores, while the slightly lower $N$ also performs quite well. This precision vs. recall trade-off is also reflected in the choice of edge threshold, as discussed in Appendix \ref{app:edge_threshold}.

\begin{table}[!ht]
    \caption{Performance of our method across different values of query budget $B$, across all settings. $\pm$ indicates the pooled standard deviation across settings. We find a budget in the range of $N - 2N$ performs best overall.}
    \label{tab:query_budget}
    \centering
    \scalebox{0.8}{
    \begin{tabular}{lccccc}
    \toprule
    Metric & $0.25N$ & $0.5N$ & $N$ & $2N$ & $\infty$ \\
    \midrule
    BSF & 0.40 $\pm$ 0.04 & 0.42 $\pm$ 0.04 & 0.48 $\pm$ 0.05 & 0.50 $\pm$ 0.06 & 0.51 $\pm$ 0.06 \\
    F1 & 0.50 $\pm$ 0.05 & 0.52 $\pm$ 0.05 & 0.57 $\pm$ 0.05 & 0.57 $\pm$ 0.05 & 0.56 $\pm$ 0.05 \\
    Recall & 0.40 $\pm$ 0.04 & 0.43 $\pm$ 0.04 & 0.50 $\pm$ 0.04 & 0.56 $\pm$ 0.05 & 0.59 $\pm$ 0.05 \\
    Precision & 0.84 $\pm$ 0.09 & 0.83 $\pm$ 0.08 & 0.77 $\pm$ 0.09 & 0.66 $\pm$ 0.08 & 0.61 $\pm$ 0.07 \\
    \bottomrule
    \end{tabular}}
\end{table}

\section{Limitations}
\label{sec:limitations}

Our work has several important limitations. Our method provides no formal guarantees of recovering the true causal graph under ideal settings. We instead aim to combine the strengths of diverse causal discovery algorithms with an imperfect expert under plausible real-world data conditions, which may violate the assumptions of individual algorithms. We must therefore rely on empirical validation.
Furthermore, our experiments involve only synthetic, categorical data. While our model averaging approach is general and should be applicable to other types of data (e.g. continuous or mixed), provided appropriate choices are made for the component algorithms, this is not yet verified empirically. Finally, while we include experiments on noisy data, the synthetic noise we use cannot be expected to fully cover all real-world conditions.

As an ensembling approach, our method inherits the worst computational complexity of any component algorithm. If considering only the model averaging, the worst-case complexity is $\mathcal{O}\!\left(N^3\right)$, where $N$ is the number of nodes, as every potential edge must be considered ($N^2$) along with a cyclicity check ($N$). Here we assume the cyclicity check is optimized with an almost-linear time algorithm for incremental graphs \citep{chen2024almost}, although we do not include this optimization in our current implementation as we find that the runtime in practice is dominated by the latency of querying an expert (e.g. an LLM). The expected runtime is therefore effectively $\Theta\!\left(L*B\right)$, where $L$ is the average query latency and $B$ is the query budget. In turn, the runtime of the ensembling process is dominated by the component algorithms.

\section{Conclusion}

We provide a flexible approach for ensembling in causal discovery using dynamically requested expert knowledge. We distinguish between edge existence and orientation and selectively query an (imperfect) expert to resolve disagreement between algorithms.
Experiments demonstrate the improvements of our approach over both ensembles without expert knowledge and individual algorithms using dynamically requested knowledge, on both clean and noisy data. 
To our knowledge, this study constitutes the first method for leveraging expert knowledge in the ensembling of causal discovery algorithms.

Future work can address certain limitations mentioned in Section \ref{sec:limitations} by employing our method on different data types and noise distributions, including on real data. Furthermore, while the current approach effectively gives every model equal weight during averaging, it may be possible to adapt the weights given to different models, based on the predicted performance of different individual algorithms. Additional possibilities include using our model averaging method across multiple runs of the same chosen algorithm rather than across an ensemble of disparate algorithms, providing improved stability (for non-deterministic algorithms or algorithms sensitive to variable ordering) or efficiency (across different data partitions). Finally, while to our knowledge this work is the first to explore the use of dynamic expert knowledge for model averaging, we foresee great potential for future exploration along this avenue.

\clearpage

\begin{ack}
This research received funding from the Flanders AI Research Program. We thank Henri Arno for his comments and suggestions.
\end{ack}

\bibliographystyle{unsrtnat}
\bibliography{references}

\appendix

\section{Data and code availability}\label{app:availability}

All of our data comes from publicly available sources: 
\begin{itemize}
    \item the Bayesys repository \citep{constantinou2020bayesys}, available at \url{https://bayesian-ai.eecs.qmul.ac.uk/bayesys/} and accessed under the CC BY-SA 4.0 license.
    \item the SimSUM dataset \citep{rabaey2025simsum}, available at \url{https://huggingface.co/datasets/prabaey/simsum} and accessed under the CC BY 4.0 license.
\end{itemize}

We further make use of the following code and packages:
\begin{itemize}
    \item pgmpy 1.0.0 \citep{Ankan2024}, MIT license
    \item causal-learn 0.1.4.3 \citep{zheng2024causal}, MIT license
    \item bnlearn 5.1 \citep{scutari2019package}, CC BY-SA 3.0 license
    \item py-tetrad \citep{ramsey2023py}, MIT license
    \item PuLP 3.3.0 \citep{mitchell2011pulp}, MIT license
    \item cdt 0.6.0 \citep{kalainathan2019causal}, MIT license
    \item DAG-GNN original code \citep{yu2019dag}, Apache 2.0 license
\end{itemize}

All data and code were used in accordance with their respective terms of use for
academic research purposes. No additional permissions were required for their use in this work.

Our code will be made available on GitHub following peer review. An anonymized version is available at \codeurl.

\section{Assumptions}

Below we discuss standard assumptions in causal discovery and provide more detail on several of our assumptions, including empirical analysis.

\subsection{Standard causal discovery assumptions}\label{app:standard_assumptions}

Common assumptions in causal discovery include:

\paragraph{Causal Markov condition.}
Let $G^\star$ be the ground-truth DAG over variables $\mathcal{V}$, and let $P$ denote the joint distribution over $\mathcal{V}$. The distribution $P$ satisfies the Causal Markov condition with respect to $G^\star$ if every $d$-separation in $G^\star$ implies a corresponding conditional independence in $P$:
\[
V_i \perp\!\!\!\perp_{G^\star} V_j \mid S \;\; \Longrightarrow \;\; V_i \perp\!\!\!\perp_P V_j \mid S,
\]
for all disjoint $V_i, V_j \in \mathcal{V}$ and $S \subseteq \mathcal{V} \setminus \{V_i, V_j\}$.The Causal Markov condition provides that $d$-separations in the causal graph are reflected by independencies in the observed data. 

\paragraph{Faithfulness.}
The distribution $P$ is faithful to $G^\star$ if every conditional independence in $P$ corresponds to a $d$-separations in $G^\star$. Formally:
\[
V_i \perp\!\!\!\perp_P V_j \mid S \;\; \Longrightarrow \;\; V_i \perp\!\!\!\perp_{G^\star} V_j \mid S,
\]
for all disjoint $V_i, V_j \in \mathcal{V}$ and $S \subseteq \mathcal{V} \setminus \{V_i, V_j\}$. Faithfulness provides that independencies in the observed data are reflected by $d$-separations in the causal graph.

\paragraph{Observability (causal sufficiency).}
All common causes of the observed variables are themselves observed - i.e., there are no latent confounders. 

\paragraph{Acyclicity.}
The ground-truth causal graph $G^\star$ is a directed acyclic graph (DAG), i.e., it contains no directed cycles.

Under these assumptions, many existing algorithms can be shown to recover the CPDAG corresponding to the Markov equivalence class of the ground-truth DAG in the infinite data limit.

\subsection{Algorithm competence}\label{app:algorithm_competence}

Condorcet's jury theorem \citep{de1785essais} provides that if a group votes on a binary outcome such that one outcome is correct and each voter has some independent probability $p > \frac{1}{2}$ of voting correctly, the probability of the correct outcome being the result of a majority vote approaches infinity as the group increases in size. Treating the existence of a given edge as a binary classification problem, the result trivially extends to the full graph. However, in practice the votes of the algorithms in the ensemble cannot be expected to be independent, as the algorithms receive the same observational data as input and operate under similar principles (typically conditional independence tests and/or a likelihood-based scoring function). The effectiveness of majority voting diminishes as the correlation between votes rises \citep{ladha1995information}, suggesting as diverse a group of algorithms as possible (e.g. constraint-based, score-based, and hybrid methods) should be used as the basis of the ensemble.

Given some correlation will doubtless remain regardless of the diversity in chosen algorithms, the performance of the ensemble will not improve arbitrarily with the addition of further algorithms. The relative value of adding further algorithms will diminish after a certain point, depending on the degree of correlation; how this behaves in practice may be an interesting topic for future research.

In a sense, the use of an expert diversifies the ensemble far more than simply adding additional data-driven algorithms can be expected to. Furthermore, the expert may make our method robust to minor violations of Assumption 1 (algorithm competence). Intuitively, if an algorithm that performs worse than random may nonetheless uncover some true edges not otherwise detected by a majority vote of the ensemble. Our method provides a path for these true edges to be added to the final model, provided the expert approves their existence.

\subsection{Expert correctness}\label{app:expert_correctness}

\begin{figure}
    \centering
    \includegraphics[width=.8\textwidth]{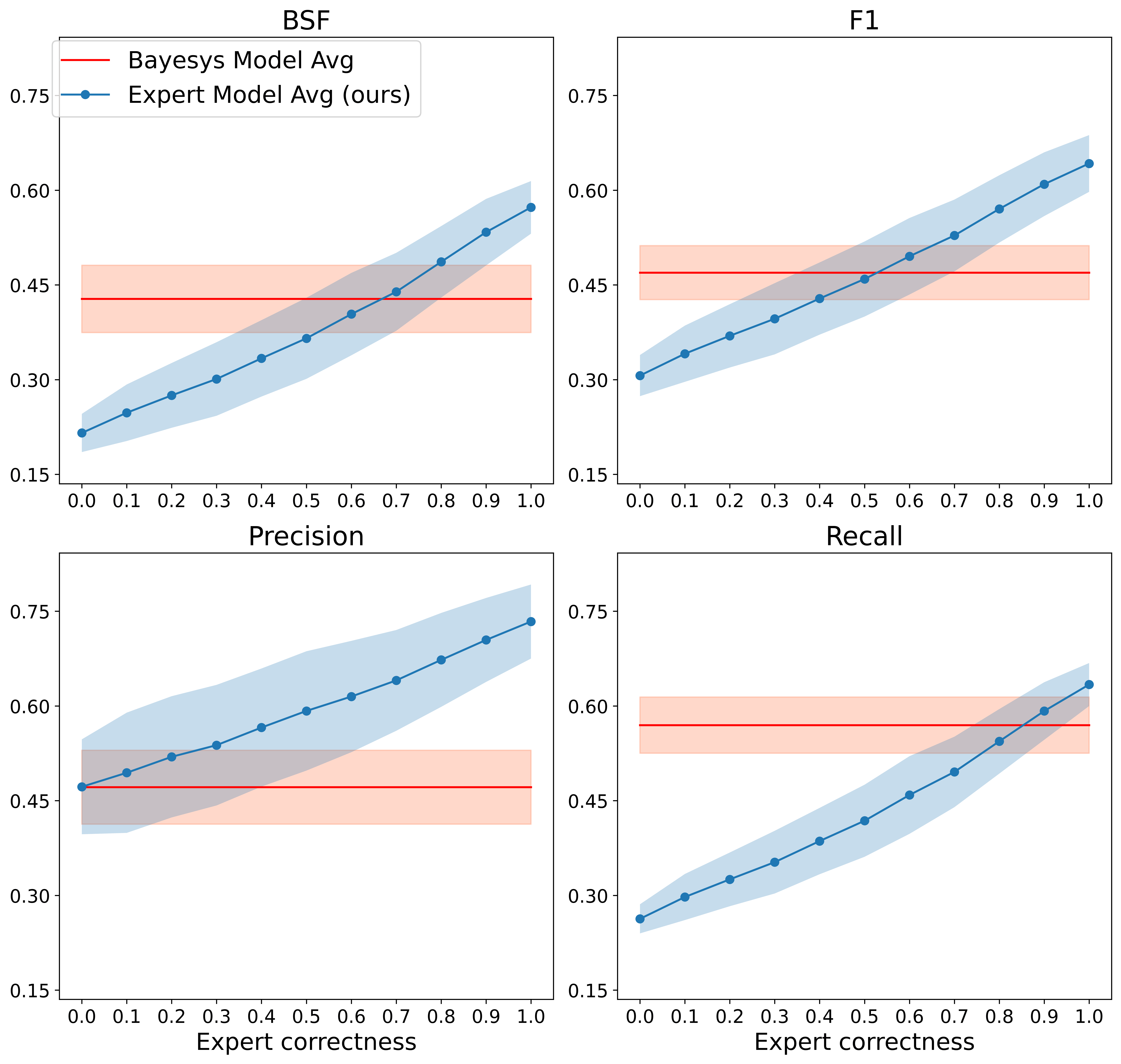}
    \caption{Performance of our method at different values of the $p_e^{\text{exist}},\; p_e^{\text{orient}}$, compared to Bayesys Model Avg as a static baseline, averaged across all settings (networks, sample sizes, and noise). Performance of our method increases linearly with expert correctness, generally surpassing performance of a method that does not use an expert at around $p_e^{\text{exist}} = 50\%,\; p_e^{\text{orient}} = 50\%$. The shaded region indicates the pooled standard deviation across settings.}
    \label{fig:expert_correctness}
\end{figure}

Figure \ref{fig:expert_correctness} shows the performance of different values of $p_e^{\text{exist}},\; p_e^{\text{orient}}$ with our method, keeping other settings fixed ($\theta_1=0.0,\;\theta_2=0.9,\;B=2N$). Performance increases linearly with expert correctness. This follows naturally: of the set of considered edges $\mathcal{E}$, some subset $\mathcal{E}_{val}$ are sent to the expert. Expected performance on $\mathcal{E}_{val}$ increases linearly with the probability of the expert providing a correct response, while performance on edges not sent to the expert is unaffected and therefore a constant factor.

Our method surpasses Bayesys Model Avg on F1 score at $p_e^{\text{exist}} \approx 50\%,\; p_e^{\text{orient}} \approx 50\%$, suggesting the use of an expert in the ensembling process is only valuable when the expert performs better than random. Even at low expert correctness, our method achieves higher precision than Bayesys Model Avg; vice-versa, even at relatively high expert correctness (up to around $80\%$), 
Bayesys Model Avg achieves higher recall than our method. This likely follows from the use of majority voting in our ensembling process: edges that pass a majority vote are added automatically, bypassing the incorrect expert. Following the exhaustion of the query budget, all decisions are made by majority voting.

\subsection{Assumption 5 validation}\label{app:assumption_validation}

To test Assumption 5 (query value), we alter our method to also query the expert in areas of low disagreement. Specifically, rather than only query the expert for edges that do not pass a majority vote, we query the expert for all edges. Tables \ref{tab:assumption_validation_clean} and \ref{tab:assumption_validation_noisy} demonstrate that our method, which prioritizes querying to resolve disagreement among algorithms, reliably improves over this naive version on all metrics save precision, at all tested query budgets. We note that our strategy of querying in areas of disagreement improves not just with lower query budgets (where the expert queries may be entirely exhausted on low-disagreement edges), but also with a high or even infinite budget. We posit that this is due to the imperfect nature of our expert: given the expert does not perfectly answer existence queries, querying the expert for edges that are very likely to be correct (presumably ones for which agreement is high) may effectively only give the expert an opportunity to incorrectly reject true edges. 

\begin{table}
\caption{The fraction of clean data settings where our method improves over a query strategy that does not prioritize areas of disagreement for queries, as determined by a one-sided Wilcoxon signed-rank test over 20 seeds ($p < 0.05$). All metrics other than precision typically improve. A * further indicates a significant improvement when considering all settings together.}
\label{tab:assumption_validation_clean}
	\centering
	\resizebox{\textwidth}{!}{
    \begin{tabular}{lcccccc}
    \toprule
    Algorithm & BSF & F1 & Precision & Recall & Relative SHD & Relative SID \\
    \midrule
    $0.25N$ & 64\% (18/28)* & 64\% (18/28)* & 29\% (8/28)* & 64\% (18/28)* & 54\% (15/28)* & 54\% (15/28)* \\
    $0.5N$ & 82\% (23/28)* & 82\% (23/28)* & 25\% (7/28) & 82\% (23/28)* & 75\% (21/28)* & 64\% (18/28)* \\
    $1N$ & 100\% (28/28)* & 96\% (27/28)* & 11\% (3/28) & 100\% (28/28)* & 68\% (19/28)* & 82\% (23/28)* \\
    $2N$ & 100\% (28/28)* & 93\% (26/28)* & 4\% (1/28) & 100\% (28/28)* & 46\% (13/28)* & 71\% (20/28)* \\
    $\infty$ & 100\% (28/28)* & 96\% (27/28)* & 54\% (15/28)* & 100\% (28/28)* & 68\% (19/28)* & 82\% (23/28)* \\
    \bottomrule
    \end{tabular}
    }
\end{table}

\begin{table}
\caption{The fraction of noisy data settings where our method improves over a query strategy that does not prioritize areas of disagreement for queries, as determined by a one-sided Wilcoxon signed-rank test over 20 seeds ($p < 0.05$). All metrics other than precision typically improve. A * further indicates a significant improvement when considering all settings together.}
\label{tab:assumption_validation_noisy}
	\centering
	\resizebox{\textwidth}{!}{
    \begin{tabular}{lcccccc}
    \toprule
    Algorithm & BSF & F1 & Precision & Recall & Relative SHD & Relative SID \\
    \midrule
    $0.25N$ & 79\% (22/28)* & 79\% (22/28)* & 18\% (5/28) & 79\% (22/28)* & 68\% (19/28)* & 71\% (20/28)* \\
    $0.5N$ & 71\% (20/28)* & 71\% (20/28)* & 18\% (5/28) & 79\% (22/28)* & 61\% (17/28)* & 68\% (19/28)* \\
    $1N$ & 86\% (24/28)* & 79\% (22/28)* & 18\% (5/28) & 86\% (24/28)* & 61\% (17/28)* & 71\% (20/28)* \\
    $2N$ & 71\% (20/28)* & 64\% (18/28)* & 29\% (8/28) & 79\% (22/28)* & 32\% (9/28)* & 68\% (19/28)* \\
    $\infty$ & 75\% (21/28)* & 71\% (20/28)* & 36\% (10/28)* & 79\% (22/28)* & 50\% (14/28)* & 68\% (19/28)* \\
    \bottomrule
    \end{tabular}
    }
\end{table}

\section{Hyperparameter and implementation details}\label{app:h_imp}

We list the implementations and hyperparameter choices used for the baseline algorithms in Table \ref{tab:h_imp}. More detail on the code packages used is available in Appendix \ref{app:availability}.

\begin{table}[ht]
    \caption{Algorithm implementations and hyperparameter choices. We typically use the hyperparameter defaults of the implementation or the recommendations of the authors, as available. \newline $\alpha$ is the dependency threshold [with dependency score], MID is the maximum in-degree}
    \label{tab:h_imp}    
    \centering
    \begin{tabular}{lp{0.35\linewidth}p{0.4\linewidth}}
        \hline
        \textbf{Algorithm} & \textbf{Software} & \textbf{Hyperparameter settings} \\
        \hline
        PC-Stable   & pgmpy \citep{Ankan2024}   & $\alpha = 0.01 [G^2]$ \\
        FCI         & causal-learn \citep{zheng2024causal}   & $\alpha = 0.01 [Chi^2]$ \\
        MMHC        & bnlearn \citep{scutari2019package}  & $\alpha = 0.05$ [MI], score BIC \\
        Tabu        & Implemented in Python 3.12.3, using: \newline \phantom{tt} pgmpy \citep{Ankan2024} & escapes = 10, score BIC \\
        GES         & pgmpy \citep{Ankan2024}  & score BIC \\
        MAHC        & Implemented in Python 3.12.3, using: \newline \phantom{tt} pgmpy \citep{Ankan2024} & MID(pruning) = 3, MID(learning) = 8, \newline score BIC \\
        BOSS        & py-tetrad \citep{ramsey2023py} & num starts = 1, use BES = False, \newline score Bdeu \\
        DAG-GNN     & original Python source code \citep{yu2019dag} & threshold = $0.3$, $\lambda=0, c=1, \tau=0$\\
        Bayesys model average & Implemented in Python 3.12.3, using: \newline \phantom{tt} pgmpy \citep{Ankan2024} & minimum count = 1\\
        Aslani-Mohebbi ensemble & Implemented in Python 3.12.3, using: \newline \phantom{tt} pgmpy \citep{Ankan2024} \newline \phantom{tt} PuLP \citep{mitchell2011pulp} & $p=0.33, q=0.15$ \\
        Tabu-AL     & Implemented in Python 3.12.3, using: \newline \phantom{tt} pgmpy \citep{Ankan2024} & escapes = 10, score BIC, low delta threshold = 0.20, small counts threshold = 0.01, unreliable score threshold = 0.001, request limit = $2N$ \\
        \hline
    \end{tabular}
\end{table}

\section{Expert model averaging pseudocode}\label{app:pseudocode}

Algorithm \ref{alg:ensemble} provides pseudocode for our model averaging strategy, described earlier in Section \ref{sec:ema_method}. Our pseudocode makes use of several helper functions, notably: \texttt{connectionCounts($\mathcal{G}$)}, which counts the total number of direct edges of any kind between each pair of nodes in a set of models $\mathcal{G}$; \texttt{edgeCounts($\mathcal{G}, x, y$)}, which counts the total number of edges $x\rightarrow y$ in a set of models $\mathcal{G}$; \texttt{$\hat{G}$.edgeCreatesCycle($x,y$)}, which determines if adding a new edge $x\rightarrow y$ to the model $\hat{G}$ would create a cycle; and \texttt{$e$.acceptConnection($x,y$)} and \texttt{$e$.determineOrientation($x,y$)}, which dynamically query an expert $e$ for edge existence and orientation.

\RestyleAlgo{ruled}

\begin{algorithm}[ht]
\caption{Expert Model Averaging}
\label{alg:ensemble}
\KwIn{a set of causal models $\mathcal{G}$, edge threshold $\theta_1$, orientation threshold $\theta_2$, query budget $B$, and expert $e$}
\KwOut{$\hat{G}$, the final model}
$C\leftarrow$ \texttt{connectionCounts($\mathcal{G}$)}\;
$n \leftarrow$ size($\mathcal{G}$)\;
$\hat{G} \leftarrow$ DAG()\;
$q \leftarrow 0$\;
\For{$x$, $y$, $c$ \kwin \text{\upshape sorted($C$)}}{
  \uIf{$c/n \ge \theta_1$ \kwand ($c/n > 0.5$ \kwor ($B > q$ \kwand $e$\text{\upshape.\texttt{acceptConnection}($x, y$)))}}{
    \uIf{$c/n \le 0.5$}{$q \mathrel{+}= 1$\;}
    $xy \leftarrow$ \kwnot  $\hat{G}$.\texttt{edgeCreatesCycle}($x, y$)\;
    $yx \leftarrow$ \kwnot  $\hat{G}$.\texttt{edgeCreatesCycle}($y, x$)\;
    \uIf{$xy$ \kwand $yx$}{
        \uIf{\text{\upshape \texttt{edgeCounts}($\mathcal{G}, x, y$)}$/c \ge \theta_2$}{
            $\hat{G}$.addEdge($x, y$)\;
        }
        \uElseIf{\text{\upshape \texttt{edgeCounts}($\mathcal{G}, y, x$)}$/c \ge \theta_2$}{
            $\hat{G}$.addEdge($y, x$)\;
        }
        \ElseIf{$B > q$}{
            $\hat{G}$.addEdge(\phantom{a}$e$.\texttt{determineOrientation}($x, y$)\phantom{a})\;
            $q \mathrel{+}= 1$\;
        }
    }
    \uElseIf{$xy$}{
        $\hat{G}$.addEdge($x, y$)\;
    }
    \ElseIf{$yx$}{
        $\hat{G}$.addEdge($y, x$)\;
    }
  }
}
\Return $\hat{G}$\;
\end{algorithm}

\section{Selection of hyperparameters}\label{app:hyperparameter_choices}

We tune the hyperparameters of our method on randomly generated networks. Using the \texttt{random.graph()} method of bnlearn \citep{scutari2019package}, we generate 15 network structures of between 5-50 nodes: three each of sizes $\{5, 10, 20, 35, 50\}$. We use a maximum in-degree of 5 and a generation method randomly chosen from among $\{\texttt{ordered, ic-dag, melancon}\}$. We generate 100000 clean samples from each network, and follow the strategy of \citet{constantinou2021large} to obtain corresponding noisy data. We then select subsets of size $n \in \{100, 1000, 10000, 100000\}$ to form the clean and noisy training sets. The generated networks and training data are available with the rest of our code at \codeurl.

Keeping the correctness of our simulated expert fixed at $p_e^{\text{exist}} = 70\%,\, p_e^{\text{orient}} = 90\%$, we test our method on the generated networks with all combinations of edge threshold $\theta_1\in\{0.0, 0.1, \dots, 0.5\}$, orientation threshold $\theta_2\in\{0.5, 0.6, \ldots, 1.0\}$, and query budget $B \in\{0.25N,0.5N,N,2N,\infty\}$. 

We choose the settings $\theta_1=0.0,\,\theta_2=0.9,\,B=2N$ as they provide the highest BSF and F1 scores, averaged across all networks and sample sizes.

\section{LLM case study}\label{app:llm_results}

\subsection{Setup}

In addition to a simulated expert, we conduct experiments with an LLM expert, using OpenAI's gpt-5-nano-2025-08-07 for all experiments. While we average all results across 20 seeds (consistent with our other experiments), calls to the LLM were cached across seeds and sample sizes for cost saving purposes, such that only one call to each of \texttt{acceptConnection} and \texttt{determineOrientation} is made for the same variables. To reduce ordering sensitivity, we query the LLM with both possible argument orders, and use the LLM's choice only if it is consistent across both orders; otherwise, we fall back to a majority vote among the ensemble models $\mathcal{G}$. We use our standard hyperparameter settings: $\theta_1=0.0,\;\theta_2=0.9,\;B=2N$.

We exclude two networks from the case study: the Pathfinder network due to the lack of semantically meaningful variable names, and the Formed network due to a high number of diagnostic rather than causal edges (e.g. \texttt{Violence} $\to$ \texttt{Age}, \texttt{ViolentConvictions} $\to$ \texttt{Violence}). We also test only the clean versions of the networks. The structure of the noisy versions of the networks differs only by the inclusion of latent confounders, which are not directly considered in the current prompts, as we believe handling latent confounders with expert input deserves targeted experiment design---as such, it poses an interesting avenue for future work.

\subsection{Results}

Table \ref{tab:llm_accuracy} shows the accuracy of LLM responses on existence and orientation queries. (Here, we assess the individual LLM responses for each variable ordering, not the results of any resulting fallbacks to majority voting.) For all networks, performance is higher on orientation queries than existence queries, indicating that the LLM is better at determining plausible causal orderings than the existence of causal relationships, supporting the results of prior research \citep{kiciman2023causal,carro2024large}.

The mean accuracy across networks is $93.3\%$ for orientation queries and $66.4\%$ for existence queries, supporting our choices of $p_e^{\text{exist}} = 70\%,\; p_e^{\text{orient}} = 90\%$ for the simulated expert. The LLM performs notably poorly on existence queries for Alarm and Property ($p_e^{\text{exist}} = 52.2\%$ and $p_e^{\text{exist}} = 43.9\%$, respectively), even violating Assumption 3 (expert competence) for on the Property network. While a more powerful model and/or more sophisticated prompting strategies may improve this performance, we find that at least some of the degradation in performance can be explained by the structure of the networks, which contain modeling choices that make it difficult for an LLM to accurately determine the existence of causal relationships between certain variables. For example, in the Alarm network the variables \texttt{KINKEDTUBE} and \texttt{SHUNT} are not in a causal relationship, although an LLM may reasonably expect a kinked tube to be a possible (indirect) cause of a pulmonary shunt by impacting ventilation. Such issues also hold for other pairs of variables (e.g. \texttt{VENTLUNG} and \texttt{SHUNT} in Alarm, \texttt{propertyExpensesGrowth} and \texttt{propertyExpenses} in Property).

\begin{table}
    \caption{LLM accuracy on orientation and existence queries for benchmark networks. Mean performance across networks is in-line with the choices of $p_e^{\text{exist}} = 70\%,\; p_e^{\text{orient}} = 90\%$ used for the simulated expert.}
    \label{tab:llm_accuracy}
    \centering
    \begin{tabular}{llll}
        \toprule
        \textbf{Network} & \textbf{Orientation} & \textbf{Existence} \\
        \midrule
        Asia     & $100.0\%$  & $68.4\%$  \\
        Sports   & $100.0\%$  & $91.7\%$  \\
        Property & $83.9\%$   & $43.9\%$  \\
        Alarm    & $82.8\%$   & $52.2\%$  \\
        SimSUM   & $100.0\%$  & $76.0\%$  \\
        \midrule
        Mean     & $93.3\%$   & $66.4\%$  \\
        \bottomrule
    \end{tabular}
\end{table}

Table \ref{tab:llm_vs_sim} directly compares the performance of the LLM vs. the simulated expert  (with $p_e^{\text{exist}} = 70\%,\; p_e^{\text{orient}} = 90\%$). Results are reasonably consistent across settings, with the LLM typically displaying greater recall but lower precision than the simulated expert. As expected from the per-query accuracies reported in Table \ref{tab:llm_accuracy}, the LLM outperforms the simulated expert by most metrics on the networks Asia, Sports, and SimSUM, and vice-versa on Property and Alarm.

\begin{table}
\caption{LLM vs simulated expert. Results are reasonably consistent across settings, with the LLM typically displaying greater recall but lower precision than the simulated expert. $\pm$ indicates one standard deviation.}
\label{tab:llm_vs_sim}
	\centering
	\resizebox{\textwidth}{!}{
	\begin{tabular}{lllllllll}
		\toprule
		\textbf{Network} & \textbf{Sample size} & \textbf{Expert} & \textbf{BSF} & \textbf{F1} & \textbf{Precision} & \textbf{Recall} & \textbf{SHD} & \textbf{SID} \\ 
		\midrule
		asia & 100 & Sim & $0.47\pm0.10$ & $0.61\pm0.06$ & $0.57\pm0.06$ & $0.67\pm0.09$  & $6.68\pm1.09$  & $16.35\pm3.54$\\
		 & 100 & LLM & $0.44\pm0.13$ & $0.60\pm0.07$ & $0.49\pm0.07$ & $0.75\pm0.07$  & $8.28\pm1.74$  & $10.65\pm2.06$\\
		 & 1000 & Sim & $0.72\pm0.09$ & $0.77\pm0.06$ & $0.67\pm0.07$ & $0.91\pm0.06$  & $4.50\pm1.37$  & $6.15\pm4.32$\\
		 & 1000 & LLM & $0.69\pm0.03$ & $0.72\pm0.02$ & $0.57\pm0.02$ & $1.00\pm0.01$  & $6.12\pm0.46$  & $0.35\pm1.57$\\
		 & 10000 & Sim & $0.70\pm0.05$ & $0.74\pm0.04$ & $0.62\pm0.06$ & $0.95\pm0.05$  & $5.28\pm1.12$  & $5.90\pm5.41$\\
		 & 10000 & LLM & $0.60\pm0.08$ & $0.67\pm0.04$ & $0.50\pm0.05$ & $1.00\pm0.00$  & $8.05\pm1.50$  & $0.00\pm0.00$\\
		 & 100000 & Sim & $0.85\pm0.07$ & $0.87\pm0.06$ & $0.80\pm0.10$ & $0.95\pm0.04$  & $2.42\pm1.28$  & $5.10\pm4.84$\\
		 & 100000 & LLM & $0.81\pm0.11$ & $0.81\pm0.09$ & $0.69\pm0.12$ & $1.00\pm0.00$  & $3.90\pm2.13$  & $0.00\pm0.00$\\
		sports & 100 & Sim & $0.28\pm0.09$ & $0.48\pm0.08$ & $0.80\pm0.14$ & $0.35\pm0.07$  & $11.18\pm1.63$  & $59.20\pm5.55$\\
		 & 100 & LLM & $0.38\pm0.09$ & $0.59\pm0.06$ & $0.79\pm0.11$ & $0.48\pm0.05$  & $9.80\pm1.63$  & $52.30\pm6.02$\\
		 & 1000 & Sim & $0.28\pm0.09$ & $0.48\pm0.08$ & $0.80\pm0.14$ & $0.35\pm0.07$  & $11.18\pm1.63$  & $59.20\pm5.55$\\
		 & 1000 & LLM & $0.38\pm0.09$ & $0.59\pm0.06$ & $0.79\pm0.11$ & $0.48\pm0.05$  & $9.80\pm1.63$  & $52.30\pm6.02$\\
		 & 10000 & Sim & $0.77\pm0.08$ & $0.86\pm0.05$ & $0.87\pm0.04$ & $0.85\pm0.08$  & $4.03\pm1.36$  & $19.90\pm11.36$\\
		 & 10000 & LLM & $0.84\pm0.02$ & $0.90\pm0.01$ & $0.82\pm0.02$ & $1.00\pm0.00$  & $3.35\pm0.49$  & $0.00\pm0.00$\\
		 & 100000 & Sim & $0.86\pm0.08$ & $0.92\pm0.05$ & $0.88\pm0.07$ & $0.96\pm0.03$  & $2.65\pm1.50$  & $11.05\pm9.46$\\
		 & 100000 & LLM & $0.84\pm0.04$ & $0.90\pm0.02$ & $0.81\pm0.04$ & $1.00\pm0.00$  & $3.45\pm0.83$  & $0.00\pm0.00$\\
		property & 100 & Sim & $0.58\pm0.04$ & $0.64\pm0.04$ & $0.70\pm0.06$ & $0.60\pm0.04$  & $20.65\pm2.29$  & $54.70\pm15.95$\\
		 & 100 & LLM & $0.67\pm0.02$ & $0.65\pm0.02$ & $0.61\pm0.04$ & $0.71\pm0.02$  & $23.42\pm2.12$  & $44.35\pm5.57$\\
		 & 1000 & Sim & $0.71\pm0.04$ & $0.75\pm0.04$ & $0.76\pm0.06$ & $0.74\pm0.04$  & $15.48\pm2.78$  & $40.05\pm16.63$\\
		 & 1000 & LLM & $0.75\pm0.03$ & $0.71\pm0.03$ & $0.64\pm0.04$ & $0.80\pm0.03$  & $20.48\pm2.58$  & $34.60\pm10.75$\\
		 & 10000 & Sim & $0.86\pm0.05$ & $0.83\pm0.03$ & $0.78\pm0.05$ & $0.88\pm0.05$  & $11.45\pm2.36$  & $27.45\pm15.01$\\
		 & 10000 & LLM & $0.85\pm0.03$ & $0.76\pm0.03$ & $0.66\pm0.04$ & $0.89\pm0.03$  & $17.50\pm2.66$  & $32.50\pm6.35$\\
		 & 100000 & Sim & $0.90\pm0.04$ & $0.87\pm0.04$ & $0.83\pm0.05$ & $0.91\pm0.04$  & $8.62\pm2.41$  & $19.05\pm13.17$\\
		 & 100000 & LLM & $0.86\pm0.02$ & $0.78\pm0.02$ & $0.69\pm0.02$ & $0.90\pm0.02$  & $15.65\pm1.57$  & $32.00\pm7.34$\\
		alarm & 100 & Sim & $0.52\pm0.03$ & $0.57\pm0.03$ & $0.59\pm0.04$ & $0.55\pm0.03$  & $38.55\pm3.47$  & $266.95\pm58.50$\\
		 & 100 & LLM & $0.50\pm0.03$ & $0.49\pm0.03$ & $0.45\pm0.03$ & $0.54\pm0.03$  & $51.45\pm3.17$  & $285.45\pm36.29$\\
		 & 1000 & Sim & $0.80\pm0.03$ & $0.78\pm0.03$ & $0.74\pm0.05$ & $0.83\pm0.03$  & $21.72\pm3.56$  & $139.55\pm42.93$\\
		 & 1000 & LLM & $0.75\pm0.01$ & $0.69\pm0.01$ & $0.61\pm0.01$ & $0.79\pm0.01$  & $33.08\pm1.44$  & $187.55\pm23.89$\\
		 & 10000 & Sim & $0.85\pm0.03$ & $0.80\pm0.03$ & $0.74\pm0.05$ & $0.88\pm0.02$  & $19.75\pm3.80$  & $131.30\pm51.94$\\
		 & 10000 & LLM & $0.83\pm0.02$ & $0.73\pm0.03$ & $0.63\pm0.03$ & $0.87\pm0.02$  & $29.50\pm3.44$  & $159.65\pm30.36$\\
		 & 100000 & Sim & $0.90\pm0.02$ & $0.82\pm0.02$ & $0.74\pm0.04$ & $0.93\pm0.02$  & $18.80\pm2.96$  & $107.15\pm57.99$\\
		 & 100000 & LLM & $0.87\pm0.01$ & $0.74\pm0.02$ & $0.62\pm0.02$ & $0.91\pm0.01$  & $29.98\pm2.51$  & $168.70\pm19.22$\\
		simsum & 100 & Sim & $0.34\pm0.04$ & $0.52\pm0.03$ & $0.71\pm0.05$ & $0.41\pm0.04$  & $26.62\pm1.72$  & $113.40\pm16.17$\\
		 & 100 & LLM & $0.40\pm0.03$ & $0.57\pm0.03$ & $0.68\pm0.04$ & $0.49\pm0.03$  & $25.85\pm1.90$  & $97.55\pm8.66$\\
		 & 1000 & Sim & $0.34\pm0.04$ & $0.52\pm0.03$ & $0.71\pm0.05$ & $0.41\pm0.04$  & $26.62\pm1.72$  & $113.40\pm16.17$\\
		 & 1000 & LLM & $0.40\pm0.03$ & $0.57\pm0.03$ & $0.68\pm0.04$ & $0.49\pm0.03$  & $25.85\pm1.90$  & $97.55\pm8.66$\\
		 & 10000 & Sim & $0.59\pm0.05$ & $0.73\pm0.04$ & $0.92\pm0.05$ & $0.61\pm0.04$  & $15.45\pm2.24$  & $103.90\pm16.56$\\
		 & 10000 & LLM & $0.67\pm0.04$ & $0.79\pm0.03$ & $0.89\pm0.03$ & $0.70\pm0.03$  & $13.38\pm1.74$  & $84.85\pm14.69$\\
		 & 100000 & Sim & $0.74\pm0.03$ & $0.84\pm0.03$ & $0.96\pm0.03$ & $0.75\pm0.03$  & $9.72\pm1.50$  & $75.40\pm19.74$\\
		 & 100000 & LLM & $0.79\pm0.01$ & $0.88\pm0.01$ & $0.98\pm0.02$ & $0.80\pm0.00$  & $7.65\pm0.67$  & $39.95\pm0.22$\\
		\bottomrule
	\end{tabular}}
\end{table}

\subsection{Prompts}

We provide our prompt templates below. The relevant substitutions (e.g. network and variable descriptions) are included in our code, available at \codeurl.
We use the following prompt for orientation queries: \\

\texttt{
We are currently constructing a causal graph for a dataset covering <dataset description>. You are an <expertise description>. Assume that there is a plausible causal relationship between the following two variables (possibly indirect). Your task is to determine the most plausible *causal ordering* between them. Causation may be direct or indirect (via unobserved or observed mediators).\\
The two variables under consideration are:\\
''<X>' with possible values <X values>, described as <X description>\\
''<Y>' with possible values <Y values>, described as <Y description>\\
Before answering, consider:\\
- Which variable is more plausibly upstream.\\
- Whether intervening on one would reasonably be expected to change the other.\\
- Whether one variable represents a more downstream outcome. Note that aggregated variables are downstream of (and directly caused by) what they aggregate.\\
- Whether either variable primarily serves as evidence of underlying processes rather than a driver.\\
A: '<X>' is a cause (possibly indirect) of '<Y>' (<X> ->* <Y>).\\
B: '<Y>' is a cause (possibly indirect) of '<X>' (<Y> ->* <X>).\\
Provide a brief causal analysis (2-4 sentences), then conclude with:\\
"The correct choice is: <A/B>" \\
}

And the following for existence queries: \\

\texttt{
We are currently constructing a causal graph for a dataset covering <dataset description>. You are an <expertise description>. Your task is to determine whether there is a plausible *causal relationship* between two variables. The relationship may be direct or indirect, and may involve unobserved mediators. You are not asked to determine direction at this stage.\\
The two variables under consideration are:\\
''<X>' with possible values <X values>, described as <X description>\\
''<Y>' with possible values <Y values>, described as <Y description>\\
Assume that:\\
- The true causal graph may include unobserved variables.\\
- Observed associations alone do not determine causal ordering.\\ \\
Before answering, consider:\\
- Whether the variables are part of the same underlying (domain-relevant) mechanism.\\
- Whether there exists a specific, domain-supported causal mechanism by which changing one would plausibly alter the other, beyond general shared factors.\\
- Whether any observed association is more likely explained by confounding or selection effects.\\
- Whether the variables operate at compatible levels (e.g., trait vs symptom, background vs outcome).\\
- Whether the variables are better understood as parallel consequences of broader factors rather than causally linked to each other.\\ \\
A: There is a plausible causal relationship between '<X>' and '<Y>'.\\
B: There is no meaningful causal relationship between '<X>' and '<Y>'; any association is likely due to shared causes or noise.\\
Provide a brief causal analysis (2-4 sentences), then conclude with:
"The correct choice is: <A/B>" \\
}

For the Alarm network, we include the following additional text in the prompts for orientation queries: \\

\texttt{
Additionally, note that:\\
- Variables that represent errors or system failures are root causes of other measurements and not downstream of anything.\\
- Measurements, indicators, or monitoring outputs depend on underlying physiological states and cannot cause them.\\
}

Likewise, we include the following additional text in the prompts for existence queries: \\

\texttt{
Choose A only if at least one of the following clearly applies:\\
- one variable directly produces the other through a known mechanism\\
- one variable is an upstream structural or pathological cause of the other\\
- one variable is a downstream aggregate or consequence of the other
}

\subsection{Memorization assessment}

We acknowledge the potential that the LLM has seen, and possibly memorized, the benchmark networks in its training data. We conduct the following case study as a limited assessment of the degree to which this has occurred, prompting the chosen LLM (\texttt{gpt-5-nano}) to provide the full structure of the networks: \\

\texttt{Provide the structure of the <network name> bayesian network as a list of edges. Format the answer as a python list of tuples strings, i.e. [(`x', `y'), ...], where (`x', `y') indicates an edge x -> y in the network. Do your best even if you don't believe you have the network memorized. Answer only with the edge list.\\}

To avoid ambiguity in the networks, we use the following substitutions for \texttt{<network name>}: `ASIA', `SPORTS Asian handicap football betting', `PROPERTY UK real estate investment', `ALARM clinical monitoring', and `SimSUM patient encounters for respiratory diseases'.

The response for Asia is almost perfect, merely eliding the aggregate variable \texttt{either}, indicating a high degree of memorization:

\texttt{[(`Asia', `Tuberculosis'), (`Smoking', `Lung Cancer'), (`Tuberculosis', `XRay'), (`Lung Cancer', `XRay'), (`Tuberculosis', `Dyspnea'), (`Lung Cancer', `Dyspnea'), (`Bronchitis', `Dyspnea'), (`Smoking', `Bronchitis')]}

The response for Sports contains 13 nodes to the network's 9, of which only three are plausibly shared: \texttt{H\_Goals} and \texttt{HTgoals}, \texttt{A\_Goals} and \texttt{ATgoals}, and \texttt{Outcome} and \texttt{HDA}. The LLM response and ground-truth network both contain 15 edges, of which only two are shared.

\texttt{[(`HomeTeam', `H\_Str'), (`AwayTeam', `A\_Str'), (`HB\_RForm', `H\_Goals'), (`AB\_RForm', `A\_Goals'), (`H\_Str', `H\_Goals'), (`A\_Str', `A\_Goals'), (`H\_Goals', `Outcome'), (`A\_Goals', `Outcome'), (`AsianHandicap', `Outcome'), (`AsianHandicap', `Odds\_Home'), (`AsianHandicap', `Odds\_Draw'), (`AsianHandicap', `Odds\_Away'), (`Odds\_Home', `Outcome'), (`Odds\_Draw', `Outcome'), (`Odds\_Away', `Outcome')]}

The response for Property contains 25 nodes compared to the network's 27, of which only two are in common (after changing snake case to camel case): \texttt{capitalGrowth} and \texttt{interestRate}. No edges are shared between the LLM response and ground truth network.

\texttt{[(`GDP\_growth', `inflation'), (`GDP\_growth', `market\_sentiment'), (`GDP\_growth', `property\_price'), (`GDP\_growth', `demand'), (`GDP\_growth', `liquidity'), (`inflation', `interest\_rate'), (`inflation', `maintenance\_cost'), (`inflation', `rent\_price'), (`unemployment', `demand'), (`unemployment', `market\_sentiment'), (`interest\_rate', `mortgage\_rate'), (`interest\_rate', `property\_price'), (`mortgage\_rate', `cash\_flow'), (`mortgage\_rate', `demand'), (`mortgage\_rate', `property\_price'), (`property\_price', `cap\_rate'), (`rent\_price', `cash\_flow'), (`occupancy\_rate', `rental\_yield'), (`occupancy\_rate', `cash\_flow'), (`demand', `property\_price'), (`supply', `property\_price'), (`location\_quality', `property\_price'), (`location\_quality', `occupancy\_rate'), (`location\_quality', `demand'), (`proximity\_transport', `location\_quality'), (`taxes', `cash\_flow'), (`taxes', `net\_present\_value'), (`regulatory\_changes', `taxes'), (`regulatory\_changes', `investment\_decision'), (`maintenance\_cost', `cash\_flow'), (`market\_sentiment', `investment\_decision'), (`risk', `investment\_decision'), (`liquidity', `investment\_decision'), (`diversification', `investment\_decision'), (`capital\_growth', `net\_present\_value'), (`cash\_flow', `net\_present\_value'), (`net\_present\_value', `investment\_decision')]}

The LLM refused to provide a response for Alarm, even after adding the text from "Do your best..." on (not initially included in the prompt), answering with an empty list.

The following response was provided for SimSUM. It contains 22 edges compared to SimSUM's 35, and 20 nodes compared to SimSUM's 16, with only 5 of the 20 (25$\%$) directly corresponding to nodes in SimSUM.

\texttt{[(`age', `pneumonia'), (`smoking', `pneumonia'), (`exposure', `pneumonia'), (`age', `asthma\_exacerbation'), (`sex', `asthma\_exacerbation'), (`smoking', `copd\_exacerbation'), (`acute\_bronchitis', `cough'), (`pneumonia', `cough'), (`pneumonia', `fever'), (`pneumonia', `sob'), (`fever', `cbc\_wbc'), (`cbc\_wbc', `pneumonia\_diagnosis'), (`cxr\_result', `pneumonia\_diagnosis'), (`pneumonia\_diagnosis', `antibiotics'), (`pneumonia\_diagnosis', `bronchodilators'), (`asthma\_exacerbation', `wheeze'), (`wheeze', `sob'), (`sob', `pulse\_ox'), (`resp\_rate', `pulse\_ox'), (`pulse\_ox', `disposition'), (`antibiotics', `disposition'), (`bronchodilators', `disposition')]}

Overall, these results point to high memorization for the Asia network, but low or none for other networks.

\section{Edge threshold experiment}\label{app:edge_threshold}

\begin{figure}
    \centering
    \includegraphics[width=.8\textwidth]{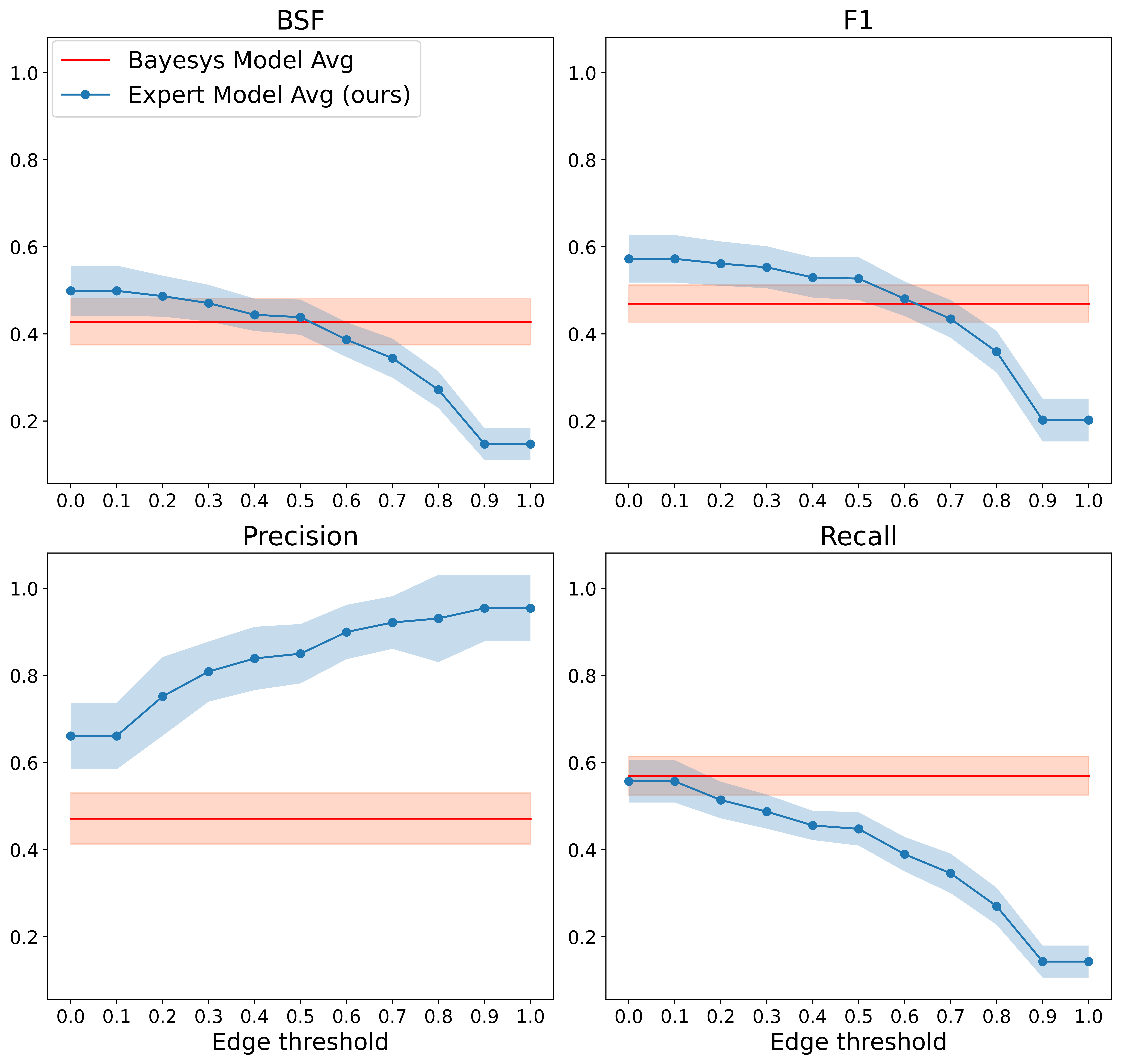}
    \caption{Performance of our method at different values of the edge threshold $\theta_1$, compared to Bayesys Model Avg as a static baseline, averaged across all settings (networks, sample sizes, and noise). As the threshold increases, precision increases at the cost of recall and BSF + F1 scores. The shaded region indicates the pooled standard deviation across settings.}
    \label{fig:edge_threshold}
\end{figure}

Figure \ref{fig:edge_threshold} shows the performance of different values of the edge threshold $\theta_1$ with our method, keeping other settings fixed ($\theta_2=0.9,\; B=2N,\; p_e^{\text{exist}} = 70\%,\; p_e^{\text{orient}} = 90\%$). As expected, precision increases along with the edge threshold, at the cost of recall. As the BSF and F1 score also decrease with higher edge thresholds, we recommend the value $\theta_1 = 0.0$ as default.

\section{Orientation threshold experiment}\label{app:orientation_threshold}

\begin{figure}
    \centering
    \includegraphics[width=.8\textwidth]{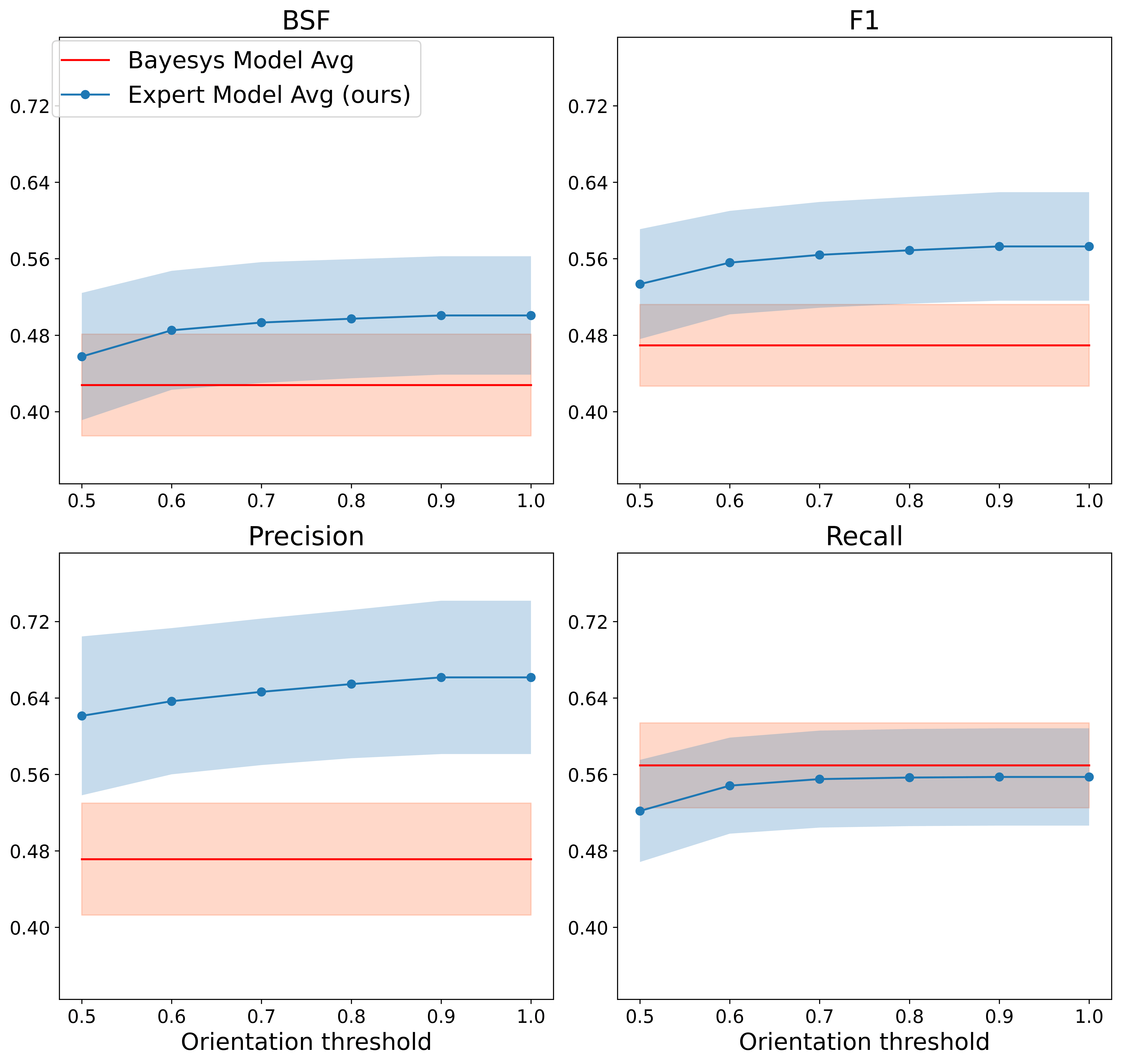}
    \caption{Performance of our method at different values of the orientation threshold $\theta_2$, compared to Bayesys Model Avg as a static baseline, averaged across all settings (networks, sample sizes, and noise). Given the high accuracy of the expert in answering orientation queries, performance typically increases as the threshold increases, as more orientation queries are sent to the expert. The shaded region indicates the pooled standard deviation across settings.}
    \label{fig:orientation_threshold}
\end{figure}

Figure \ref{fig:orientation_threshold} shows the performance of different values of the orientation threshold $\theta_2$ with our method, keeping other settings fixed ($\theta_1=0.0,\; B=2N,\; p_e^{\text{exist}} = 70\%,\; p_e^{\text{orient}} = 90\%$). Unlike with the edge threshold, there is no precision/recall trade-off---nor should one be expected, as the orientation queries do not gate edge inclusion. Instead, given the high accuracy of the expert in answering orientation queries, performance across all metrics typically increases as the threshold increases, as more orientation queries are sent to the expert.

\section{Direct edge expert}\label{app:direct_edge_expert}

As noted in Section \ref{sec:sim_expert}, we assume the simulated expert cannot distinguish between a direct edge and a chain when determining if an edge between two variables should be permitted. Compared to an expert with unrealistic access to knowledge of direct edges, this should be expected to harm precision but have no effect on recall: the expert without knowledge of direct edges will only allow more incorrect edges into the final graph, not block new edges. Tables \ref{tab:direct_clean} and \ref{tab:direct_noisy} examine how often an expert with knowledge of direct edges performs significantly better than the simulated expert we use in our other experiments. As expected, the difference is more often significant with larger query budgets, as the expert is called for more edges. Furthermore, there is typically no significant change in recall. The SID also remains mostly unchanged, as the additional spurious edges allowed by the default expert are still consistent with the true causal ordering.

\begin{table}
\caption{The fraction of clean data settings where an expert with access to direct edges significantly improves over our default expert, as determined by a one-sided Wilcoxon signed-rank test over 20 seeds ($p < 0.05$). Precision often improves but recall is unaffected. A * further indicates a significant improvement when considering all settings together.}
\label{tab:direct_clean}
	\centering
	\resizebox{\textwidth}{!}{
    \begin{tabular}{lcccccc}
    \toprule
    Query budget & BSF & F1 & Precision & Recall & Relative SHD & Relative SID \\
    \midrule
    $0.25N$ & 0\% (0/28) & 0\% (0/28) & 0\% (0/28) & 0\% (0/28) & 0\% (0/28) & 0\% (0/28) \\
    $0.5N$ & 4\% (1/28)* & 4\% (1/28)* & 4\% (1/28) & 0\% (0/28) & 4\% (1/28)* & 4\% (1/28) \\
    $N$ & 25\% (7/28)* & 36\% (10/28)* & 43\% (12/28)* & 4\% (1/28) & 39\% (11/28)* & 11\% (3/28)* \\
    $2N$ & 57\% (16/28)* & 75\% (21/28)* & 89\% (25/28)* & 4\% (1/28) & 86\% (24/28)* & 14\% (4/28)* \\
    $\infty$ & 79\% (22/28)* & 93\% (26/28)* & 93\% (26/28)* & 7\% (2/28) & 89\% (25/28)* & 18\% (5/28)* \\
    \bottomrule
    \end{tabular}}
\end{table}

\begin{table}
\caption{The fraction of noisy data settings where an expert with access to direct edges significantly improves over our default expert, as determined by a one-sided Wilcoxon signed-rank test over 20 seeds ($p < 0.05$). Precision often improves but recall is unaffected. A * further indicates a significant improvement when considering all settings together.}
\label{tab:direct_noisy}
	\centering
	\resizebox{\textwidth}{!}{
    \begin{tabular}{lcccccc}
    \toprule
    Query budget & BSF & F1 & Precision & Recall & Relative SHD & Relative SID \\
    \midrule
    $0.25N$ & 4\% (1/28) & 4\% (1/28) & 4\% (1/28) & 0\% (0/28) & 4\% (1/28) & 0\% (0/28) \\
    $0.5N$ & 7\% (2/28)* & 7\% (2/28) & 7\% (2/28) & 7\% (2/28) & 11\% (3/28)* & 7\% (2/28) \\
    $N$ & 21\% (6/28)* & 29\% (8/28)* & 36\% (10/28)* & 0\% (0/28) & 25\% (7/28)* & 7\% (2/28)* \\
    $2N$ & 39\% (11/28)* & 46\% (13/28)* & 79\% (22/28)* & 0\% (0/28) & 50\% (14/28)* & 18\% (5/28) \\
    $\infty$ & 46\% (13/28)* & 50\% (14/28)* & 79\% (22/28)* & 0\% (0/28) & 54\% (15/28)* & 18\% (5/28) \\
    \bottomrule
    \end{tabular}}
\end{table}

\section{Compute resources}\label{app:compute}

We run all experiments on internal clusters. As DAG-GNN is the only algorithm that requires a GPU, we report its resource usage separately.

Hyperparameter tuning:
\begin{itemize}
    \item Baseline algorithms: 22 hours, 20x Xeon(R) Silver 4310 CPU @ 2.10GHz, 32GB RAM
    \item DAG-GNN: 13 hours, 20x AMD EPYC 9275F 24-Core Processor, 32GB RAM, 1x NVIDIA H200 NVL
    \item Expert model average: 2 hours, 1x Xeon(R) Silver 4310 CPU @ 2.10GHz, 2GB RAM
\end{itemize}

Benchmark networks:
\begin{itemize}
    \item Baseline algorithms: 83 hours, 20x Intel(R) Xeon(R) Silver 4108 CPU @ 1.80GHz, 32GB RAM
    \item DAG-GNN: 84 hours, 4x Intel(R) Xeon(R) Silver 4108 CPU @ 1.80GHz, 32GB RAM, 1x NVIDIA GeForce RTX 4090
    \item Expert model average: computed alongside the SID (reported below)
\end{itemize}

Computing the SID:
\begin{itemize}
    \item Baseline algorithms (including DAG-GNN): 156 hours, 2x Intel(R) Xeon(R) Silver 4108 CPU @ 1.80GHz, 10GB RAM
    \item Expert model average: 4 hours, 1x Intel(R) Xeon(R) Silver 4310 CPU @ 2.10GHz, 2GB RAM
\end{itemize}

LLM case study:
\begin{itemize}
    \item 5 hours, 1x Intel(R) Xeon(R) Silver 4310 CPU @ 2.10GHz, 2GB RAM
\end{itemize}

Estimated total compute:
\begin{itemize}
    \item DAG-GNN: 30 hours, 20x AMD EPYC 9275F 24-Core Processor, 32GB RAM, 1x NVIDIA H200 NVL
    \item Other: 126 hours, 20x Xeon(R) Silver 4310 CPU @ 2.10GHz, 32GB RAM
\end{itemize}

\section{Complete results}\label{app:complete_results}

Tables \ref{tab:pc_stable_results}-\ref{tab:ema_0.0_0.9_sim_0.7_0.9_results} report the BSF, SHD, F1-score, precision, recall, and SID of methods on individual networks and training sizes, for clean and noisy data, averaged across 20 seeds. For methods that do not return a DAG, the upper and lower and bounds of the SID are reported. In all tables, A * or - indicates some or all runs, respectively, timed out ($>$6h) or yielded an exception. In the case of precision, $\dagger$ or nan indicates some or all runs, respectively, yielded an invalid precision score. $\pm$ indicates one standard deviation. A \textbf{bold} entry indicates the method was the Oracle baseline (best individual component algorithm of the ensemble) for that particular metric \& setting.

\begin{table}[t]
	\caption{PC-Stable Complete Results}
	\label{tab:pc_stable_results}
	\resizebox{\textwidth}{!}{
\begin{tabular}{llllllllll}
		\hline
		\textbf{Network} & \textbf{Samples} & \textbf{Data} & \textbf{BSF} & \textbf{F1} & \textbf{Precision} & \textbf{Recall} & \textbf{SHD} & \textbf{SID low} & \textbf{SID high} \\
		\hline
		asia & 100 & Clean & $0.01\pm0.00$ & $0.11\pm0.00$ & $0.33\pm0.00$ & $0.06\pm0.00$ & $8.50\pm0.00$ & $25.00\pm0.00$ & $35.00\pm0.00$ \\
		 & 100 & Noisy & $0.10\pm0.00$ & $0.25\pm0.00$ & $0.50\pm0.00$ & $0.17\pm0.00$ & $6.00\pm0.00$ & $\mathbf{9.00\pm0.00}$ & $23.00\pm0.00$ \\
		 & 1000 & Clean & $0.31\pm0.00$ & $0.48\pm0.00$ & $\mathbf{1.00\pm0.00}$ & $0.31\pm0.00$ & $5.50\pm0.00$ & $22.00\pm0.00$ & $41.00\pm0.00$ \\
		 & 1000 & Noisy & $0.18\pm0.00$ & $0.35\pm0.00$ & $0.60\pm0.00$ & $0.25\pm0.00$ & $5.50\pm0.00$ & $\mathbf{5.00\pm0.00}$ & $24.00\pm0.00$ \\
		 & 10000 & Clean & $0.75\pm0.00$ & $0.86\pm0.00$ & $\mathbf{1.00\pm0.00}$ & $0.75\pm0.00$ & $2.00\pm0.00$ & $2.00\pm0.00$ & $9.00\pm0.00$ \\
		 & 10000 & Noisy & $0.45\pm0.00$ & $0.61\pm0.00$ & $0.64\pm0.00$ & $0.58\pm0.00$ & $4.50\pm0.00$ & $11.00\pm0.00$ & $\mathbf{13.00\pm0.00}$ \\
		 & 100000 & Clean & $0.81\pm0.00$ & $0.90\pm0.00$ & $\mathbf{1.00\pm0.00}$ & $0.81\pm0.00$ & $1.50\pm0.00$ & $\mathbf{0.00\pm0.00}$ & $12.00\pm0.00$ \\
\hline
	\end{tabular}}
\end{table}

\begin{table}[t]
	\caption{FCI Complete Results}
	\label{tab:fci_results}
	\resizebox{\textwidth}{!}{
	\begin{tabular}{llllllllll}
		\hline
		\textbf{Network} & \textbf{Samples} & \textbf{Data} & \textbf{BSF} & \textbf{F1} & \textbf{Precision} & \textbf{Recall} & \textbf{SHD} & \textbf{SID low} & \textbf{SID high} \\
		\hline
		asia & 100 & Clean & $0.19\pm0.00$ & $0.32\pm0.00$ & $\mathbf{1.00\pm0.00}$ & $0.19\pm0.00$ & $6.50\pm0.00$ & $17.00\pm0.00$ & $36.00\pm0.00$ \\
		 & 100 & Noisy & $0.02\pm0.00$ & $0.13\pm0.00$ & $0.33\pm0.00$ & $0.08\pm0.00$ & $6.50\pm0.00$ & $12.00\pm0.00$ & $19.00\pm0.00$ \\
		 & 1000 & Clean & $0.38\pm0.00$ & $0.55\pm0.00$ & $1.00\pm0.00$ & $0.38\pm0.00$ & $5.00\pm0.00$ & $21.00\pm0.00$ & $45.00\pm0.00$ \\
		 & 1000 & Noisy & $0.18\pm0.00$ & $0.35\pm0.00$ & $0.60\pm0.00$ & $0.25\pm0.00$ & $5.50\pm0.00$ & $5.00\pm0.00$ & $24.00\pm0.00$ \\
		 & 10000 & Clean & $0.44\pm0.00$ & $0.61\pm0.00$ & $1.00\pm0.00$ & $0.44\pm0.00$ & $4.50\pm0.00$ & $13.00\pm0.00$ & $36.00\pm0.00$ \\
		 & 10000 & Noisy & $0.15\pm0.00$ & $0.40\pm0.00$ & $0.38\pm0.00$ & $0.42\pm0.00$ & $7.50\pm0.00$ & $8.00\pm0.00$ & $27.00\pm0.00$ \\
		 & 100000 & Clean & $0.50\pm0.00$ & $0.67\pm0.00$ & $1.00\pm0.00$ & $0.50\pm0.00$ & $4.00\pm0.00$ & $13.00\pm0.00$ & $28.00\pm0.00$ \\
\hline
	\end{tabular}}
\end{table}

\begin{table}[t]
	\caption{MMHC Complete Results}
	\label{tab:mmhc_results}
	\resizebox{\textwidth}{!}{
	\begin{tabular}{lllllllll}
		\hline
		\textbf{Network} & \textbf{Samples} & \textbf{Data} & \textbf{BSF} & \textbf{F1} & \textbf{Precision} & \textbf{Recall} & \textbf{SHD} & \textbf{SID}\\
		\hline
		asia & 100 & Clean & $0.26\pm0.04$ & $0.41\pm0.05$ & $1.00\pm0.00$ & $0.26\pm0.04$ & $5.92\pm0.34$ & $34.05\pm3.97$ \\
		 & 100 & Noisy & $\mathbf{0.20\pm0.05}$ & $0.34\pm0.07$ & $\mathbf{1.00\pm0.00}$ & $0.20\pm0.05$ & $\mathbf{4.78\pm0.30}$ & $19.85\pm4.23$ \\
		 & 1000 & Clean & $0.48\pm0.09$ & $0.65\pm0.08$ & $1.00\pm0.00$ & $0.48\pm0.09$ & $\mathbf{4.12\pm0.70}$ & $28.00\pm9.69$ \\
		 & 1000 & Noisy & $\mathbf{0.34\pm0.05}$ & $\mathbf{0.50\pm0.06}$ & $\mathbf{1.00\pm0.00}$ & $\mathbf{0.34\pm0.05}$ & $\mathbf{3.98\pm0.30}$ & $17.85\pm3.92$ \\
		 & 10000 & Clean & $0.57\pm0.10$ & $0.72\pm0.09$ & $1.00\pm0.00$ & $0.57\pm0.10$ & $3.48\pm0.82$ & $30.50\pm10.90$ \\
		 & 10000 & Noisy & $0.38\pm0.08$ & $0.55\pm0.06$ & $0.60\pm0.03$ & $0.51\pm0.08$ & $4.95\pm0.46$ & $22.35\pm4.66$ \\
		 & 100000 & Clean & $0.57\pm0.10$ & $0.72\pm0.09$ & $1.00\pm0.00$ & $0.57\pm0.10$ & $3.48\pm0.82$ & $30.50\pm10.90$ \\
\hline
	\end{tabular}}
\end{table}

\begin{table}[t]
	\caption{BOSS Complete Results}
	\label{tab:boss_results}
	\resizebox{\textwidth}{!}{
	\begin{tabular}{llllllllll}
		\hline
		\textbf{Network} & \textbf{Samples} & \textbf{Data} & \textbf{BSF} & \textbf{F1} & \textbf{Precision} & \textbf{Recall} & \textbf{SHD} & \textbf{SID low} & \textbf{SID high} \\
		\hline
		asia & 100 & Clean & $0.04\pm0.12$ & $0.37\pm0.07$ & $0.31\pm0.05$ & $\mathbf{0.47\pm0.10}$ & $12.75\pm1.28$ & $\mathbf{11.70\pm3.63}$ & $39.05\pm3.03$ \\
		 & 100 & Noisy & $0.20\pm0.00$ & $\mathbf{0.40\pm0.00}$ & $0.50\pm0.00$ & $\mathbf{0.33\pm0.00}$ & $6.00\pm0.00$ & $13.00\pm0.00$ & $31.00\pm0.00$ \\
		 & 1000 & Clean & $\mathbf{0.65\pm0.09}$ & $\mathbf{0.71\pm0.05}$ & $0.58\pm0.04$ & $\mathbf{0.91\pm0.06}$ & $5.90\pm1.01$ & $\mathbf{1.10\pm3.75}$ & $\mathbf{6.55\pm4.36}$ \\
		 & 1000 & Noisy & $0.05\pm0.00$ & $0.29\pm0.00$ & $0.33\pm0.00$ & $0.25\pm0.00$ & $7.50\pm0.00$ & $15.00\pm0.00$ & $29.00\pm0.00$ \\
		 & 10000 & Clean & $0.68\pm0.00$ & $0.74\pm0.00$ & $0.64\pm0.00$ & $0.88\pm0.00$ & $5.00\pm0.00$ & $\mathbf{0.00\pm0.00}$ & $12.00\pm0.00$ \\
		 & 10000 & Noisy & $0.40\pm0.20$ & $0.56\pm0.15$ & $0.56\pm0.12$ & $0.57\pm0.17$ & $5.15\pm1.53$ & $\mathbf{5.45\pm0.51}$ & $15.90\pm9.19$ \\
		 & 100000 & Clean & $0.73\pm0.09$ & $0.82\pm0.07$ & $0.84\pm0.07$ & $0.79\pm0.06$ & $2.85\pm1.08$ & $0.00\pm0.00$ & $14.40\pm7.39$ \\
\hline
	\end{tabular}}
\end{table}

\begin{table}[t]
	\caption{Tabu Complete Results}
	\label{tab:tabu_results}
	\resizebox{\textwidth}{!}{
	\begin{tabular}{lllllllll}
		\hline
		\textbf{Network} & \textbf{Samples} & \textbf{Data} & \textbf{BSF} & \textbf{F1} & \textbf{Precision} & \textbf{Recall} & \textbf{SHD} & \textbf{SID}\\
		\hline
		asia & 100 & Clean & $\mathbf{0.39\pm0.00}$ & $\mathbf{0.56\pm0.00}$ & $0.78\pm0.00$ & $0.44\pm0.00$ & $\mathbf{5.50\pm0.00}$ & $26.00\pm0.00$ \\
		 & 100 & Noisy & $0.20\pm0.07$ & $0.33\pm0.09$ & $1.00\pm0.00$ & $0.20\pm0.07$ & $4.80\pm0.41$ & $20.20\pm5.75$ \\
		 & 1000 & Clean & $0.40\pm0.01$ & $0.57\pm0.01$ & $0.67\pm0.01$ & $0.50\pm0.01$ & $5.98\pm0.11$ & $37.05\pm2.44$ \\
		 & 1000 & Noisy & $0.33\pm0.00$ & $0.50\pm0.00$ & $1.00\pm0.00$ & $0.33\pm0.00$ & $4.00\pm0.00$ & $17.10\pm0.45$ \\
		 & 10000 & Clean & $\mathbf{0.86\pm0.06}$ & $\mathbf{0.89\pm0.03}$ & $0.88\pm0.01$ & $\mathbf{0.91\pm0.06}$ & $\mathbf{1.70\pm0.47}$ & $\mathbf{8.40\pm5.64}$ \\
		 & 10000 & Noisy & $0.45\pm0.00$ & $0.61\pm0.00$ & $0.64\pm0.00$ & $0.58\pm0.00$ & $4.50\pm0.00$ & $17.95\pm0.22$ \\
		 & 100000 & Clean & $\mathbf{0.88\pm0.01}$ & $\mathbf{0.94\pm0.01}$ & $1.00\pm0.00$ & $\mathbf{0.88\pm0.01}$ & $\mathbf{0.98\pm0.11}$ & $\mathbf{11.75\pm1.12}$ \\
\hline
	\end{tabular}}
\end{table}

\begin{table}[t]
	\caption{GES Complete Results}
	\label{tab:ges_results}
	\resizebox{\textwidth}{!}{
	\begin{tabular}{lllllllll}
		\hline
		\textbf{Network} & \textbf{Samples} & \textbf{Data} & \textbf{BSF} & \textbf{F1} & \textbf{Precision} & \textbf{Recall} & \textbf{SHD} & \textbf{SID}\\
		\hline
		asia & 100 & Clean & $0.35\pm0.06$ & $0.52\pm0.06$ & $0.76\pm0.03$ & $0.40\pm0.06$ & $5.80\pm0.47$ & $30.80\pm7.52$ \\
		 & 100 & Noisy & $0.17\pm0.00$ & $0.29\pm0.00$ & $1.00\pm0.00$ & $0.17\pm0.00$ & $5.00\pm0.00$ & $23.00\pm0.00$ \\
		 & 1000 & Clean & $0.40\pm0.00$ & $0.57\pm0.00$ & $0.67\pm0.00$ & $0.50\pm0.00$ & $6.00\pm0.00$ & $36.00\pm0.00$ \\
		 & 1000 & Noisy & $0.33\pm0.00$ & $0.50\pm0.00$ & $1.00\pm0.00$ & $0.33\pm0.00$ & $4.00\pm0.00$ & $\mathbf{17.00\pm0.00}$ \\
		 & 10000 & Clean & $0.60\pm0.15$ & $0.72\pm0.11$ & $0.77\pm0.09$ & $0.68\pm0.13$ & $4.15\pm1.53$ & $32.70\pm13.65$ \\
		 & 10000 & Noisy & $\mathbf{0.47\pm0.12}$ & $\mathbf{0.62\pm0.09}$ & $0.64\pm0.05$ & $\mathbf{0.60\pm0.12}$ & $\mathbf{4.38\pm0.72}$ & $16.20\pm3.35$ \\
		 & 100000 & Clean & $0.52\pm0.20$ & $0.67\pm0.15$ & $0.68\pm0.18$ & $0.67\pm0.12$ & $5.55\pm2.76$ & $29.60\pm13.16$ \\
\hline
	\end{tabular}}
\end{table}

\begin{table}[t]
	\caption{MAHC Complete Results}
	\label{tab:mahc_results}
	\resizebox{\textwidth}{!}{
	\begin{tabular}{lllllllll}
		\hline
		\textbf{Network} & \textbf{Samples} & \textbf{Data} & \textbf{BSF} & \textbf{F1} & \textbf{Precision} & \textbf{Recall} & \textbf{SHD} & \textbf{SID}\\
		\hline
		asia & 100 & Clean & $0.25\pm0.00$ & $0.40\pm0.00$ & $1.00\pm0.00$ & $0.25\pm0.00$ & $6.00\pm0.00$ & $34.00\pm0.00$ \\
		 & 100 & Noisy & $0.00\pm0.00$ & $0.00\pm0.00$ & nan & $0.00\pm0.00$ & $6.00\pm0.00$ & $\mathbf{18.00\pm0.00}$ \\
		 & 1000 & Clean & $0.38\pm0.00$ & $0.55\pm0.00$ & $1.00\pm0.00$ & $0.38\pm0.00$ & $5.00\pm0.00$ & $27.00\pm0.00$ \\
		 & 1000 & Noisy & $0.33\pm0.00$ & $0.50\pm0.00$ & $1.00\pm0.00$ & $0.33\pm0.00$ & $4.00\pm0.00$ & $17.00\pm0.00$ \\
		 & 10000 & Clean & $0.69\pm0.00$ & $0.81\pm0.00$ & $1.00\pm0.00$ & $0.69\pm0.00$ & $2.50\pm0.00$ & $33.00\pm0.00$ \\
		 & 10000 & Noisy & $0.20\pm0.00$ & $0.40\pm0.00$ & $0.50\pm0.00$ & $0.33\pm0.00$ & $6.00\pm0.00$ & $22.00\pm0.00$ \\
		 & 100000 & Clean & $0.69\pm0.00$ & $0.81\pm0.00$ & $1.00\pm0.00$ & $0.69\pm0.00$ & $2.50\pm0.00$ & $33.00\pm0.00$ \\
\hline
	\end{tabular}}
\end{table}

\begin{table}[t]
	\caption{DAG-GNN Complete Results}
	\label{tab:dag_gnn_results}
	\resizebox{\textwidth}{!}{
	\begin{tabular}{llllllllll}
		\hline
		\textbf{Network} & \textbf{Samples} & \textbf{Data} & \textbf{BSF} & \textbf{F1} & \textbf{Precision} & \textbf{Recall} & \textbf{SHD} & \textbf{SID low} & \textbf{SID high} \\
		\hline
		asia & 100 & Clean & $0.16\pm0.06$ & $0.37\pm0.05$ & $0.57\pm0.28$ & $0.31\pm0.08$ & $8.57\pm1.81$ & $20.55\pm3.17$ & $\mathbf{22.85\pm3.62}$ \\
		 & 100 & Noisy & $0.01\pm0.09$ & $0.10\pm0.10$ & $0.43\pm0.41\dagger$ & $0.06\pm0.06$ & $6.40\pm0.97$ & $18.25\pm2.79$ & $18.25\pm2.79$ \\
		 & 1000 & Clean & $0.51\pm0.07$ & $0.66\pm0.06$ & $0.80\pm0.11$ & $0.57\pm0.05$ & $4.68\pm0.98$ & $25.50\pm2.31$ & $25.50\pm2.31$ \\
		 & 1000 & Noisy & $0.00\pm0.00$ & $0.00\pm0.00$ & nan & $0.00\pm0.00$ & $6.00\pm0.00$ & $18.00\pm0.00$ & $18.00\pm0.00$ \\
		 & 10000 & Clean & $0.17\pm0.14$ & $0.45\pm0.07$ & $0.42\pm0.15$ & $0.52\pm0.05$ & $10.78\pm3.05$ & $21.75\pm5.21$ & $21.75\pm5.21$ \\
		 & 10000 & Noisy & $0.02\pm0.07$ & $0.03\pm0.11$ & $\mathbf{1.00\pm0.00\dagger}$ & $0.02\pm0.07$ & $5.90\pm0.45$ & $17.55\pm2.01$ & $17.55\pm2.01$ \\
		 & 100000 & Clean & - & - & - & - & - & - & - \\
\hline
	\end{tabular}}
\end{table}

\begin{table}[t]
	\caption{Tabu-AL Complete Results}
	\label{tab:tabu_al_rl_2.0_results}
	\resizebox{\textwidth}{!}{
	\begin{tabular}{lllllllll}
		\hline
		\textbf{Network} & \textbf{Samples} & \textbf{Data} & \textbf{BSF} & \textbf{F1} & \textbf{Precision} & \textbf{Recall} & \textbf{SHD} & \textbf{SID}\\
		\hline
		asia & 100 & Clean & $0.37\pm0.20$ & $0.54\pm0.18$ & $0.77\pm0.24$ & $0.43\pm0.16$ & $5.70\pm2.16$ & $17.55\pm8.15$ \\
		 & 100 & Noisy & $0.16\pm0.14$ & $0.29\pm0.18$ & $0.72\pm0.40$ & $0.18\pm0.12$ & $5.25\pm1.12$ & $13.25\pm3.32$ \\
		 & 1000 & Clean & $0.35\pm0.24$ & $0.54\pm0.17$ & $0.57\pm0.20$ & $0.53\pm0.14$ & $7.42\pm3.26$ & $19.20\pm11.41$ \\
		 & 1000 & Noisy & $0.16\pm0.14$ & $0.29\pm0.18$ & $0.72\pm0.40$ & $0.18\pm0.12$ & $5.25\pm1.12$ & $13.25\pm3.32$ \\
		 & 10000 & Clean & $0.17\pm0.28$ & $0.46\pm0.15$ & $0.40\pm0.17$ & $0.56\pm0.14$ & $11.25\pm4.47$ & $27.00\pm11.77$ \\
		 & 10000 & Noisy & $0.34\pm0.15$ & $0.52\pm0.12$ & $0.58\pm0.14$ & $0.48\pm0.12$ & $5.25\pm1.27$ & $13.00\pm5.23$ \\
		 & 100000 & Clean & $0.08\pm0.46$ & $0.45\pm0.23$ & $0.39\pm0.24$ & $0.55\pm0.18$ & $13.00\pm7.21$ & $26.90\pm12.81$ \\
\hline
	\end{tabular}}
\end{table}

\begin{table}[t]
	\caption{Bayesys Model Avg Complete Results}
	\label{tab:bayesys_ma_results}
	\resizebox{\textwidth}{!}{
	\begin{tabular}{llllllllll}
		\hline
		\textbf{Network} & \textbf{Samples} & \textbf{Data} & \textbf{BSF} & \textbf{F1} & \textbf{Precision} & \textbf{Recall} & \textbf{SHD} & \textbf{SID low} & \textbf{SID high} \\
		\hline
		asia & 100 & Clean & $0.08\pm0.10$ & $0.42\pm0.05$ & $0.32\pm0.04$ & $0.62\pm0.10$ & $13.78\pm1.48$ & $19.85\pm3.86$ & $33.50\pm3.47$ \\
		 & 100 & Noisy & $0.18\pm0.05$ & $0.40\pm0.04$ & $0.46\pm0.06$ & $0.35\pm0.03$ & $6.38\pm0.60$ & $16.35\pm1.23$ & $29.85\pm2.21$ \\
		 & 1000 & Clean & $0.38\pm0.07$ & $0.56\pm0.04$ & $0.46\pm0.04$ & $0.72\pm0.06$ & $8.98\pm0.95$ & $20.45\pm7.39$ & $24.30\pm7.27$ \\
		 & 1000 & Noisy & $0.13\pm0.00$ & $0.36\pm0.00$ & $0.40\pm0.00$ & $0.33\pm0.00$ & $7.00\pm0.00$ & $15.00\pm0.00$ & $27.10\pm0.45$ \\
		 & 10000 & Clean & $0.31\pm0.14$ & $0.53\pm0.07$ & $0.40\pm0.07$ & $0.80\pm0.07$ & $11.48\pm2.44$ & $16.05\pm5.04$ & $16.05\pm5.04$ \\
		 & 10000 & Noisy & $0.28\pm0.04$ & $0.50\pm0.03$ & $0.44\pm0.03$ & $0.58\pm0.03$ & $7.05\pm0.54$ & $17.00\pm2.49$ & $19.75\pm2.65$ \\
		 & 100000 & Clean & $0.65\pm0.13$ & $0.73\pm0.09$ & $0.65\pm0.13$ & $0.85\pm0.04$ & $5.22\pm2.28$ & $13.10\pm2.27$ & $13.10\pm2.27$ \\
\hline
	\end{tabular}}
\end{table}

\begin{table}[t]
	\caption{Aslani \& Mohebbi Complete Results}
	\label{tab:aslani_mohebbi_results}
	\resizebox{\textwidth}{!}{
	\begin{tabular}{lllllllll}
		\hline
		\textbf{Network} & \textbf{Samples} & \textbf{Data} & \textbf{BSF} & \textbf{F1} & \textbf{Precision} & \textbf{Recall} & \textbf{SHD} & \textbf{SID}\\
		\hline
		asia & 100 & Clean & $0.09\pm0.11$ & $0.42\pm0.06$ & $0.32\pm0.04$ & $0.63\pm0.12$ & $13.68\pm1.39$ & $24.85\pm6.61$ \\
		 & 100 & Noisy & $0.22\pm0.10$ & $0.42\pm0.08$ & $0.48\pm0.08$ & $0.38\pm0.08$ & $6.18\pm0.82$ & $27.20\pm5.77$ \\
		 & 1000 & Clean & $0.35\pm0.06$ & $0.54\pm0.03$ & $0.45\pm0.03$ & $0.68\pm0.04$ & $9.18\pm0.91$ & $29.00\pm3.78$ \\
		 & 1000 & Noisy & $0.13\pm0.00$ & $0.36\pm0.00$ & $0.40\pm0.00$ & $0.33\pm0.00$ & $7.00\pm0.00$ & $20.00\pm0.00$ \\
		 & 10000 & Clean & $0.32\pm0.13$ & $0.54\pm0.06$ & $0.40\pm0.07$ & $0.81\pm0.05$ & $11.40\pm2.40$ & $15.15\pm3.75$ \\
		 & 10000 & Noisy & $0.27\pm0.04$ & $0.49\pm0.03$ & $0.43\pm0.03$ & $0.58\pm0.04$ & $7.10\pm0.50$ & $17.80\pm3.65$ \\
		 & 100000 & Clean & $0.66\pm0.12$ & $0.74\pm0.09$ & $0.68\pm0.12$ & $0.83\pm0.04$ & $4.78\pm1.99$ & $15.05\pm3.63$ \\
\hline
	\end{tabular}}
\end{table}

\begin{table}[t]
	\caption{EMA (ours) Complete Results, $B=0.25N$}
	\label{tab:ema_0.0_0.9_sim_0.7_0.9_rl_0.25_results}
	\resizebox{\textwidth}{!}{
	\begin{tabular}{lllllllll}
		\hline
		\textbf{Network} & \textbf{Samples} & \textbf{Data} & \textbf{BSF} & \textbf{F1} & \textbf{Precision} & \textbf{Recall} & \textbf{SHD} & \textbf{SID}\\
		\hline
		asia & 100 & Clean & $0.49\pm0.03$ & $0.66\pm0.03$ & $1.00\pm0.00$ & $0.49\pm0.03$ & $4.08\pm0.24$ & $25.45\pm2.82$ \\
		 & 100 & Noisy & $0.26\pm0.06$ & $0.42\pm0.06$ & $0.90\pm0.17$ & $0.28\pm0.04$ & $4.58\pm0.59$ & $12.95\pm3.19$ \\
		 & 1000 & Clean & $0.60\pm0.06$ & $0.74\pm0.05$ & $0.91\pm0.08$ & $0.62\pm0.05$ & $3.55\pm0.63$ & $27.30\pm4.28$ \\
		 & 1000 & Noisy & $0.40\pm0.03$ & $0.58\pm0.03$ & $1.00\pm0.00$ & $0.40\pm0.03$ & $3.58\pm0.18$ & $12.75\pm1.83$ \\
		 & 10000 & Clean & $0.82\pm0.05$ & $0.90\pm0.03$ & $1.00\pm0.00$ & $0.82\pm0.05$ & $1.40\pm0.42$ & $18.25\pm5.63$ \\
		 & 10000 & Noisy & $0.43\pm0.03$ & $0.60\pm0.03$ & $0.63\pm0.01$ & $0.57\pm0.03$ & $4.60\pm0.21$ & $21.00\pm2.05$ \\
		 & 100000 & Clean & $0.86\pm0.03$ & $0.92\pm0.02$ & $1.00\pm0.00$ & $0.86\pm0.03$ & $1.12\pm0.22$ & $14.50\pm2.74$ \\
\hline
	\end{tabular}}
\end{table}

\begin{table}[t]
	\caption{EMA (ours) Complete Results, $B=0.5N$}
	\label{tab:ema_0.0_0.9_sim_0.7_0.9_rl_0.5_results}
	\resizebox{\textwidth}{!}{
	\begin{tabular}{lllllllll}
		\hline
		\textbf{Network} & \textbf{Samples} & \textbf{Data} & \textbf{BSF} & \textbf{F1} & \textbf{Precision} & \textbf{Recall} & \textbf{SHD} & \textbf{SID}\\
		\hline
		asia & 100 & Clean & $0.53\pm0.03$ & $0.70\pm0.03$ & $1.00\pm0.00$ & $0.53\pm0.03$ & $3.72\pm0.26$ & $19.85\pm3.62$ \\
		 & 100 & Noisy & $0.28\pm0.07$ & $0.46\pm0.07$ & $0.75\pm0.15$ & $0.33\pm0.07$ & $4.75\pm0.55$ & $10.30\pm2.94$ \\
		 & 1000 & Clean & $0.63\pm0.06$ & $0.76\pm0.05$ & $0.91\pm0.08$ & $0.66\pm0.05$ & $3.30\pm0.70$ & $25.10\pm5.10$ \\
		 & 1000 & Noisy & $0.48\pm0.04$ & $0.65\pm0.03$ & $1.00\pm0.00$ & $0.48\pm0.04$ & $3.12\pm0.22$ & $6.45\pm2.63$ \\
		 & 10000 & Clean & $0.89\pm0.04$ & $0.94\pm0.03$ & $1.00\pm0.00$ & $0.89\pm0.04$ & $0.88\pm0.36$ & $12.70\pm5.64$ \\
		 & 10000 & Noisy & $0.51\pm0.04$ & $0.65\pm0.03$ & $0.66\pm0.01$ & $0.65\pm0.04$ & $4.12\pm0.22$ & $14.45\pm2.63$ \\
		 & 100000 & Clean & $0.91\pm0.03$ & $0.95\pm0.02$ & $1.00\pm0.00$ & $0.91\pm0.03$ & $0.72\pm0.26$ & $10.20\pm3.86$ \\
\hline
	\end{tabular}}
\end{table}

\begin{table}[t]
	\caption{EMA (ours) Complete Results, $B=N$}
	\label{tab:ema_0.0_0.9_sim_0.7_0.9_rl_1.0_results}
	\resizebox{\textwidth}{!}{
	\begin{tabular}{lllllllll}
		\hline
		\textbf{Network} & \textbf{Samples} & \textbf{Data} & \textbf{BSF} & \textbf{F1} & \textbf{Precision} & \textbf{Recall} & \textbf{SHD} & \textbf{SID}\\
		\hline
		asia & 100 & Clean & $0.49\pm0.09$ & $0.65\pm0.08$ & $0.75\pm0.12$ & $0.57\pm0.06$ & $5.08\pm1.25$ & $19.95\pm3.71$ \\
		 & 100 & Noisy & $0.43\pm0.14$ & $0.59\pm0.11$ & $0.76\pm0.15$ & $0.49\pm0.11$ & $4.05\pm1.11$ & $9.90\pm3.92$ \\
		 & 1000 & Clean & $0.84\pm0.09$ & $0.89\pm0.06$ & $0.93\pm0.07$ & $0.87\pm0.07$ & $1.62\pm0.94$ & $7.90\pm5.80$ \\
		 & 1000 & Noisy & $0.35\pm0.07$ & $0.54\pm0.06$ & $0.62\pm0.13$ & $0.48\pm0.04$ & $5.03\pm0.92$ & $6.85\pm2.41$ \\
		 & 10000 & Clean & $0.91\pm0.05$ & $0.93\pm0.04$ & $0.92\pm0.07$ & $0.95\pm0.05$ & $1.12\pm0.65$ & $6.25\pm5.68$ \\
		 & 10000 & Noisy & $0.58\pm0.06$ & $0.68\pm0.04$ & $0.61\pm0.05$ & $0.78\pm0.05$ & $4.32\pm0.63$ & $4.65\pm3.44$ \\
		 & 100000 & Clean & $0.92\pm0.05$ & $0.93\pm0.04$ & $0.91\pm0.06$ & $0.95\pm0.04$ & $1.12\pm0.63$ & $5.55\pm4.99$ \\
\hline
	\end{tabular}}
\end{table}

\begin{table}[t]
	\caption{EMA (ours) Complete Results, $B=2N$}
	\label{tab:ema_0.0_0.9_sim_0.7_0.9_rl_2.0_results}
	\resizebox{\textwidth}{!}{
	\begin{tabular}{lllllllll}
		\hline
		\textbf{Network} & \textbf{Samples} & \textbf{Data} & \textbf{BSF} & \textbf{F1} & \textbf{Precision} & \textbf{Recall} & \textbf{SHD} & \textbf{SID}\\
		\hline
		asia & 100 & Clean & $0.47\pm0.10$ & $0.61\pm0.06$ & $0.57\pm0.06$ & $0.67\pm0.09$ & $6.68\pm1.09$ & $16.35\pm3.54$ \\
		 & 100 & Noisy & $0.43\pm0.14$ & $0.60\pm0.11$ & $0.75\pm0.16$ & $0.50\pm0.11$ & $4.03\pm1.12$ & $9.45\pm3.07$ \\
		 & 1000 & Clean & $0.72\pm0.09$ & $0.77\pm0.06$ & $0.67\pm0.07$ & $0.91\pm0.06$ & $4.50\pm1.37$ & $6.15\pm4.32$ \\
		 & 1000 & Noisy & $0.35\pm0.07$ & $0.54\pm0.06$ & $0.62\pm0.13$ & $0.48\pm0.04$ & $5.03\pm0.92$ & $6.85\pm2.41$ \\
		 & 10000 & Clean & $0.70\pm0.05$ & $0.74\pm0.04$ & $0.62\pm0.06$ & $0.95\pm0.05$ & $5.28\pm1.12$ & $5.90\pm5.41$ \\
		 & 10000 & Noisy & $0.53\pm0.07$ & $0.65\pm0.05$ & $0.56\pm0.05$ & $0.78\pm0.05$ & $5.03\pm0.79$ & $4.35\pm3.25$ \\
		 & 100000 & Clean & $0.85\pm0.07$ & $0.87\pm0.06$ & $0.80\pm0.10$ & $0.95\pm0.04$ & $2.42\pm1.28$ & $5.10\pm4.84$ \\
\hline
	\end{tabular}}
\end{table}

\begin{table}[t]
	\caption{EMA (ours) Complete Results, $B=\infty$}
	\label{tab:ema_0.0_0.9_sim_0.7_0.9_results}
	\resizebox{\textwidth}{!}{
	\begin{tabular}{lllllllll}
		\hline
		\textbf{Network} & \textbf{Samples} & \textbf{Data} & \textbf{BSF} & \textbf{F1} & \textbf{Precision} & \textbf{Recall} & \textbf{SHD} & \textbf{SID}\\
		\hline
		asia & 100 & Clean & $0.42\pm0.09$ & $0.58\pm0.06$ & $0.51\pm0.06$ & $0.69\pm0.09$ & $7.88\pm1.29$ & $17.15\pm3.91$ \\
		 & 100 & Noisy & $0.43\pm0.14$ & $0.60\pm0.11$ & $0.75\pm0.16$ & $0.50\pm0.11$ & $4.03\pm1.12$ & $9.45\pm3.07$ \\
		 & 1000 & Clean & $0.72\pm0.09$ & $0.77\pm0.06$ & $0.67\pm0.07$ & $0.91\pm0.06$ & $4.50\pm1.37$ & $6.15\pm4.32$ \\
		 & 1000 & Noisy & $0.35\pm0.07$ & $0.54\pm0.06$ & $0.62\pm0.13$ & $0.48\pm0.04$ & $5.03\pm0.92$ & $6.85\pm2.41$ \\
		 & 10000 & Clean & $0.68\pm0.05$ & $0.73\pm0.04$ & $0.60\pm0.06$ & $0.95\pm0.05$ & $5.72\pm1.29$ & $5.90\pm5.41$ \\
		 & 10000 & Noisy & $0.53\pm0.07$ & $0.65\pm0.05$ & $0.56\pm0.05$ & $0.78\pm0.05$ & $5.03\pm0.79$ & $4.35\pm3.25$ \\
		 & 100000 & Clean & $0.85\pm0.07$ & $0.87\pm0.06$ & $0.80\pm0.10$ & $0.95\pm0.04$ & $2.42\pm1.28$ & $5.10\pm4.84$ \\
\hline
	\end{tabular}}
\end{table}

\section{Excluded results}\label{app:excluded_results}

A "-" in the complete result tables in Appendix \ref{app:complete_results} generally indicates that all runs of that algorithm for that network, data type, and sample size timed out ($>$6h), with the following exceptions:
\begin{itemize}
    \item Aslani \& Mohebbi ensembling ran out of memory ($>$32GB) on the formed and pathfinder networks.
    \item BOSS ran out of memory ($>$32GB) on the pathfinder network.
\end{itemize}
The only algorithm that yielded exceptions on some (but not all) seeds for certain network, data type, and sample size combinations was BOSS, which occasionally yielded an invalid comparator error.

To help quantify the range of the potential bias introduced by excluding missing results, we perform simple best-case and worst-case imputations. For settings where an algorithm returned results under some seeds but not others, we impute the best or worst result, respectively, obtained by that algorithm for that setting. For settings where an algorithm returned no results, we impute the best or worst result of algorithms that did return results for that seed. We create new versions of Tables \ref{tab:sig_clean} and \ref{tab:sig_noisy} that display the range of percentages under which our algorithm significantly improves over other methods using these imputations: Tables \ref{tab:sig_clean_impute} and \ref{tab:sig_noisy_impute}.

Given the imputation strategy, the range is typically larger for algorithms that did not obtain results for certain settings (e.g. BOSS, DAG-GNN, and Aslani \& Mohebbi). In general, as results are typically missing in more challenging settings, we expect the ``true'' value to lie closer to the worst-case imputation than the best-case. However, in practice, we find excluding the missing results often produces a value around the middle of the range provided by the imputations, which we feel supports it as a reasonable conservative default.

\begin{table}
    \caption{Percentage range of settings where our method significantly improves over baselines on clean data with missing results imputed. Parentheses indicate the percentage with missing results excluded. Significance is determined by a one-sided Wilcoxon signed-rank test over 20 seeds ($p < 0.05$), demonstrating improvement in most settings.}
    \label{tab:sig_clean_impute}
    \centering
    \resizebox{\textwidth}{!}{
    \begin{tabular}{lccccccc}
    \toprule
    Algorithm & BSF & F1 & Precision & Recall & Relative SHD & Relative SID (lower bound) & Relative SID (upper bound) \\
    \midrule
    PC-Stable & 89--93\% (92\%) & 68--71\% (69\%) & 4--11\% (4\%) & 93--100\% (100\%) & 50--54\% (50\%) & 39--46\% (42\%) & 79--86\% (85\%) \\
    FCI & 93--96\% (96\%) & 79--82\% (81\%) & 0--7\% (0\%) & 93--100\% (100\%) & 50--54\% (50\%) & 29--36\% (31\%) & 93--100\% (100\%) \\
    MMHC & 100--100\% (100\%) & 86--86\% (86\%) & 11--11\% (11\%) & 100--100\% (100\%) & 54--54\% (54\%) & 89--89\% (89\%) & 89--89\% (89\%) \\
    BOSS & 64--75\% (71\%) & 68--82\% (79\%) & 43--61\% (58\%) & 54--68\% (62\%) & 61--79\% (71\%) & 25--46\% (33\%) & 71--93\% (83\%) \\
    Tabu & 86--86\% (86\%) & 82--82\% (82\%) & 50--50\% (50\%) & 86--86\% (86\%) & 71--71\% (71\%) & 86--86\% (86\%) & 86--86\% (86\%) \\
    GES & 100--100\% (100\%) & 96--96\% (96\%) & 46--46\% (46\%) & 100--100\% (100\%) & 82--82\% (82\%) & 96--96\% (96\%) & 96--96\% (96\%) \\
    MAHC & 89--96\% (96\%) & 82--89\% (88\%) & 21--29\% (23\%) & 86--93\% (92\%) & 71--79\% (77\%) & 93--100\% (100\%) & 93--100\% (100\%) \\
    DAG-GNN & 75--93\% (91\%) & 79--96\% (95\%) & 64--89\% (82\%) & 71--89\% (86\%) & 75--93\% (91\%) & 68--89\% (86\%) & 71--89\% (86\%) \\
    Tabu-AL & 79--89\% (88\%) & 75--86\% (83\%) & 61--79\% (75\%) & 75--93\% (88\%) & 79--93\% (92\%) & 68--82\% (79\%) & 68--82\% (79\%) \\
    Bayesys Model Avg & 61--61\% (61\%) & 96--96\% (96\%) & 100--100\% (100\%) & 36--36\% (36\%) & 100--100\% (100\%) & 61--61\% (61\%) & 79--79\% (79\%) \\
    Aslani \& Mohebbi & 57--82\% (75\%) & 71--93\% (90\%) & 71--100\% (100\%) & 25--54\% (35\%) & 75--100\% (100\%) & 61--89\% (85\%) & 61--89\% (85\%) \\
    Oracle* & 50--50\% (50\%) & 57--57\% (57\%) & 0--0\% (0\%) & 54--54\% (54\%) & 18--18\% (18\%) & 4--11\% (7\%) & 57--64\% (57\%) \\
    \bottomrule
    \end{tabular}
    }
\end{table}

\begin{table}
    \caption{Percentage range of settings where our method significantly improves over baselines on noisy data with missing results imputed. Parentheses indicate the percentage with missing results excluded. Significance is determined by a one-sided Wilcoxon signed-rank test over 20 seeds ($p < 0.05$), demonstrating improvement in most settings.}
    \label{tab:sig_noisy_impute}
    \centering
    \resizebox{\textwidth}{!}{
    \begin{tabular}{lccccccc}
    \toprule
    Algorithm & BSF & F1 & Precision & Recall & Relative SHD & Relative SID (lower bound) & Relative SID (upper bound) \\
    \midrule
    PC-Stable & 96--96\% (96\%) & 93--93\% (93\%) & 21--25\% (22\%) & 96--100\% (100\%) & 61--61\% (59\%) & 46--50\% (48\%) & 86--89\% (89\%) \\
    FCI & 89--96\% (96\%) & 93--100\% (100\%) & 50--61\% (56\%) & 89--100\% (100\%) & 75--82\% (80\%) & 25--36\% (28\%) & 82--93\% (92\%) \\
    MMHC & 93--93\% (93\%) & 89--89\% (89\%) & 4--14\% (4\%) & 100--100\% (100\%) & 57--57\% (57\%) & 86--86\% (86\%) & 86--86\% (86\%) \\
    BOSS & 82--93\% (92\%) & 79--89\% (92\%) & 39--54\% (46\%) & 79--93\% (100\%) & 71--82\% (79\%) & 43--57\% (50\%) & 75--89\% (88\%) \\
    Tabu & 93--93\% (93\%) & 89--89\% (89\%) & 25--36\% (28\%) & 100--100\% (100\%) & 64--64\% (64\%) & 93--93\% (93\%) & 93--93\% (93\%) \\
    GES & 93--93\% (93\%) & 96--96\% (96\%) & 25--36\% (28\%) & 100--100\% (100\%) & 79--79\% (79\%) & 93--93\% (93\%) & 93--93\% (93\%) \\
    MAHC & 93--93\% (93\%) & 96--96\% (96\%) & 25--39\% (29\%) & 100--100\% (100\%) & 68--68\% (68\%) & 93--93\% (93\%) & 93--93\% (93\%) \\
    DAG-GNN & 75--86\% (82\%) & 86--100\% (100\%) & 46--75\% (62\%) & 86--93\% (91\%) & 64--82\% (77\%) & 64--86\% (82\%) & 79--86\% (82\%) \\
    Tabu-AL & 93--96\% (96\%) & 89--93\% (92\%) & 25--32\% (27\%) & 93--100\% (100\%) & 64--68\% (65\%) & 64--79\% (77\%) & 64--79\% (77\%) \\
    Bayesys Model Avg & 71--71\% (71\%) & 71--71\% (71\%) & 96--96\% (96\%) & 39--39\% (39\%) & 96--96\% (96\%) & 64--64\% (64\%) & 82--82\% (82\%) \\
    Aslani \& Mohebbi & 68--86\% (80\%) & 68--86\% (80\%) & 68--96\% (95\%) & 36--64\% (50\%) & 68--93\% (90\%) & 64--93\% (90\%) & 64--93\% (90\%) \\
    Oracle* & 64--64\% (64\%) & 68--68\% (68\%) & 0--0\% (0\%) & 86--86\% (93\%) & 29--29\% (29\%) & 18--18\% (18\%) & 68--68\% (68\%) \\
    \bottomrule
    \end{tabular}
    }
\end{table}


\end{document}